\documentclass{article}

\usepackage{hyperref}

\usepackage{url}            
\usepackage{booktabs}       
\usepackage{amsfonts}       
\usepackage{nicefrac}       
\usepackage{microtype}      
\usepackage{epstopdf}
\usepackage{color}
\usepackage{amsmath,amssymb,amsfonts,amsthm}
\usepackage{multirow}
\usepackage{mathtools}
\usepackage{subcaption}
\usepackage{adjustbox}

\newcommand{\ra}{\rightarrow}

\newcommand{\lb}{\left(}
\newcommand{\rb}{\right)}
\newcommand{\trace}{\textnormal{trace}}
\newcommand{\diag}{\textnormal{diag}}

\newcommand{\hH}{\widehat{H}}
\newcommand{\tH}{\widetilde{H}}

\newcommand{\pQ}{Q^{\prime}}

\newcommand{\pG}{G^{\prime}}
\newcommand{\oG}{\overline{G}}
\newcommand{\od}{\overline{d}}

\newcommand{\pc}{{c^{\prime}}}
\newcommand{\pER}{p_{\textnormal{ER}}}
\newcommand{\pWS}{p_{\textnormal{WS}}}

\newcommand{\bPhi}{\mathbf{\Phi}}

\newcommand{\bpL}{\mathbf{L}^{\prime}}

\newcommand{\bLNN}{\mathbf{L}_{N}}

\newcommand{\bpLN}{\mathbf{L}^{\prime}_{\mathcal{N}}}

\newcommand{\bW}{\mathbf{W}}

\newcommand{\bL}{\mathbf{L}}

\newcommand{\bS}{\mathbf{S}}

\newcommand{\bpW}{\mathbf{W}^{\prime}}

\newcommand{\boW}{\overline{\mathbf{W}}}

\newcommand{\cG}{\mathcal{G}}
\newcommand{\cV}{\mathcal{V}}
\newcommand{\cE}{\mathcal{E}}

\newcommand{\cpV}{\mathcal{V}^{\prime}}
\newcommand{\cpE}{\mathcal{E}^{\prime}}

\newcommand{\coE}{\mathcal{\overline{E}}}

\newtheorem{thm} {Theorem}
\newtheorem{lem} {Lemma}
\newtheorem{cor} {Corollary}

\usepackage[accepted]{icml2019}

\icmltitlerunning{Fast Incremental von Neumann Graph Entropy Computation:  Theory, Algorithm, and Applications}

\begin{document}

\twocolumn[
\icmltitle{Fast Incremental von Neumann Graph Entropy Computation:\\  Theory, Algorithm, and Applications}



\icmlsetsymbol{equal}{*}

\begin{icmlauthorlist}
\icmlauthor{Pin-Yu Chen}{IBM}
\icmlauthor{Lingfei Wu}{IBM}
\icmlauthor{Sijia Liu}{IBM}
\icmlauthor{Indika Rajapakse}{Umich}
\end{icmlauthorlist}

\icmlaffiliation{IBM}{IBM Research}
\icmlaffiliation{Umich}{University of Michigan, Ann Arbor, USA}

\icmlcorrespondingauthor{Pin-Yu Chen}{pin-yu.chen@ibm.com}

\icmlkeywords{Machine Learning, ICML}
\icmlkeywords{Network analysis, ICML}
\icmlkeywords{Graph mining, ICML}
\vskip 0.3in
]



\printAffiliationsAndNotice{}  

\begin{abstract}
The von Neumann graph entropy (VNGE) facilitates measurement of information divergence and distance between graphs in a graph sequence. It has been successfully applied to various learning tasks driven by network-based data. While effective, VNGE is computationally demanding as it requires the full eigenspectrum of the graph Laplacian matrix.  In this paper, we propose a new computational framework,
 \textbf{F}ast \textbf{I}ncremental von \textbf{N}eumann \textbf{G}raph \textbf{E}nt\textbf{R}opy (FINGER), which  approaches VNGE with a performance guarantee. FINGER reduces the cubic complexity of VNGE to linear complexity in the number of nodes and edges, and thus enables online computation based on incremental graph changes. We also show asymptotic equivalence of FINGER to the exact VNGE, and derive its approximation error bounds. Based on FINGER, we propose efficient algorithms for computing Jensen-Shannon distance between graphs. Our experimental results on different random graph models demonstrate the computational efficiency and the asymptotic equivalence of FINGER. In addition, we apply FINGER to two real-world applications and one synthesized anomaly detection dataset, and corroborate its superior performance over seven baseline graph similarity methods.
\end{abstract}

\section{Introduction}
\label{sec_Intro}
In recent years, graph-based learning has become an active research field \citep{Shuman13,kalofolias2016learn,luo2012forging,shivanna2014learning,wang2016graph,kipf2017semi,wu2018scalable,wu2018d2ke,xu2018graph2seq}. Its success is rooted in the advanced capability of summarizing and representing phenomenal structural features embedded in graphs. 
In particular, evaluating similarity between graphs is crucial to network analysis and graph-based anomaly detection \citep{papadimitriou2010web,Akoglu15Graphanomalydetection,ranshous2015anomaly}. For example, 
Yanardag and Vishwanathan used graph similarity for learning novel graph kernels \citep{yanardag2015structural}, and Sharpnack et al. proposed the Lovasz extended scan statistic for anomaly detection in connected graphs \citep{sharpnack2013near}. Koutra et al.  proposed DeltaCon, a state-of-the-art similarity algorithm in terms of its scalability and capability of handling weighted graphs
using fast belief propagation \citep{koutra2016d}. However, these methods are sensitive to heuristic metrics and presumed models, and thus provide limited understanding on the general notion of variations between graphs. On the other hand, model-agnostic approaches such as graph entropy have been used to quantify the structural complexity of a single graph, which relates to the Shannon entropy of a probability distribution over a function of enumerated subgraphs in a graph  \citep{simonyi1995graph,shetty2005discovering,li2016structural}. However, graph entropy can be computationally demanding due to its  use of exhaustive subgraph search. 

Different from the aforementioned approaches and inspired by quantum information theory, the von Neumann graph entropy (VNGE) \citep{braunstein2006laplacian,passerini2008neumann,passerini2009quantifying}
facilitates the  measure of (quantum) Jensen-Shannon divergence and distance \citep{endres2003new,briet2009properties} between graphs. It associates with a model-agnostic information measure for quantifying variation between two quantum density matrices.  In addition, the VNGE has been shown to be linearly correlated with classical graph entropy measures \citep{anand2009entropy,anand2011shannon}.
The VNGE and the Jensen-Shannon distance have been successfully applied to structural reduction in multiplex networks \citep{de2015structural}, depth analysis in image processing \citep{han2012graph,bai2014depth}, structure-function analysis in genetic networks \citep{seaman2017nucleome,liu2018dynamic}, and network-ensemble comparison \citep{li2018network}. However, 
despite its effectiveness, the computation of VNGE requires (at most) cubic complexity in the number of nodes, thereby impeding its applicability to machine learning and data mining tasks involving a sequence of large graphs.

\textbf{Contributions.} To overcome the computational inefficiency of VNGE, we propose a \textbf{F}ast \textbf{I}ncremental von \textbf{N}eumann \textbf{G}raph \textbf{E}nt\textbf{R}opy (FINGER) framework to approximate VNGE with a performance guarantee, reducing its cubic complexity to linear complexity in the number of nodes and edges.
FINGER is a generic tool that applies to both batch and online graph sequences. It enables fast entropy computation when every single graph in a graph sequence is presented (e.g., a snapshot of a dynamic network, or a single-layer connectivity pattern of a multiplex network). For applications where changes in a graph (e.g., addition and deletion of nodes or edges over time) are continuously reported (e.g., streaming graphs), FINGER also allows online computation based on incremental graph changes.
We prove that FINGER maintains an approximation guarantee and is asymptotically equivalent to  the exact VNGE under some eigenspectrum conditions, which is further validated by different synthetic random graphs. 
We then apply FINGER to developing efficient algorithms for the computation of Jensen-Shannon distance between graphs. Comparing to the state-of-the-art graph similarity methods and two alternative approximate VNGE, FINGER yields superior and robust performance for anomaly detection in evolving Wikipedia networks and router communication networks, as well as bifurcation analysis in dynamic genomic networks. 
These applications show the effectiveness and potentials of  Jensen-Shannon distance for network learning in a wide range of domains, which has not been rigorously explored owing to its high computation complexity in the absence of FINGER.

The contributions of this paper and the proposed framework (FINGER) are summarized as follows.\\
$\bullet$ Two types of approximate VNGE reducing its cubic complexity to linear complexity are proposed to support fast and incremental computation of VNGE.
We derive their approximation error bounds and show asymptotic equivalence relative to the exact VNGE under mild conditions. \\
$\bullet$  FINGER achieves nearly 100\% reduction in computation time for VNGE of 
different graphs and
enables scalable Jensen-Shannon graph distance computation. \\
$\bullet$ On two real-world applications (anomaly detection and cellular bifurcation analysis) and one synthesized dataset, 
FINGER exhibits outstanding and robust performance over 7 baseline and state-of-the-art methods.

\textbf{Related Work.} 
The VNGE was firstly defined based on the combinatorial graph Laplacian matrix \citep{braunstein2006laplacian,passerini2008neumann,passerini2009quantifying,de2015structural,li2018network}. Variants of VNGE and their approximations have been proposed in the literature, including the  normalized graph Laplacian matrix \citep{Shi00} proposed in \citep{han2012graph} and the generalized graph Laplacian matrix  of directed graphs \citep{chung2005laplacians} proposed in \citep{ye2014approximate}.
However, these alternatives lack approximation justification and are shown to be suboptimal in Section \ref{sec_app}. To the best of our knowledge, this paper is the first work that provides fast VNGE computation with a provable approximation analysis.

\section{FINGER: Theory and Algorithms}
\label{Sec_FINGER}

\subsection{Background and Preliminaries}

Using terminology from quantum statistical mechanics, a density matrix $\bPhi$ describing a quantum system in a mixed state can be cast as a statistical ensemble of several quantum states. The $n \times n$ matrix $\bPhi$ is 
symmetric, positive semidefinite, and satisfies $\trace(\bPhi)=1$. The von Neumann entropy   of a quantum system is defined as $H=-\trace(\bPhi \ln \bPhi)$  \citep{von1955mathematical}, where $\ln \bPhi$ denotes matrix logarithm. Let $\{\lambda_i\}_{i=1}^n$ be the sorted eigenvalues of $\bPhi$ such that $0 \leq \lambda_n \leq \ldots \leq \lambda_1 $. The definition of von Neumann entropy is equivalent to $H=-\sum_{i=1}^n \lambda_i \ln \lambda_i$, where the convention $0 \ln 0 = 0 $ is used due to $\lim_{x \ra 0^+ }x \ln x =0$. Moreover, since $\sum_{i}\lambda_i=1$ and $\lambda_i \geq 0$ for all $i$, the von Neumann entropy can be viewed as the Shannon entropy associated with the eigenspectrum $\{\lambda_i\}_{i=1}^n$.

We consider the class of undirected weighted simple non-empty graphs with nonnegative edge weights, denoted by $\cG$.
Let $G=(\cV,\cE,\bW) \in \cG$ denote a single graph, where $\cV$ and $\cE$ denote its node and edge set with cardinality $|\cV|=n$ and $|\cE|=m$, respectively, and $\bW$ is an $n \times n$ matrix with entry $[\bW]_{ij}=w_{ij}$ denoting the weight of an edge $(i,j) \in \cE$.  A graph sequence $\{G_t\}_{t=1}^T$  refers to a set of $T$ graphs indexed by $t \in \{1,\ldots,T\}$ with known node-to-node correspondence, where $G_t \in \cG$ for all $t$.
The combinatorial graph Laplacian matrix of $G$ is defined as $\bL=\bS-\bW$ \citep{Luxburg07}, where $\bS=\diag(s_1,\ldots,s_n)$ is a diagonal matrix and its diagonal entry $s_i=\sum_{j=1}^n w_{ij}$ is the nodal strength (weighted degree) of a node $i \in \cV$.
Connecting the von Neumann entropy to graphs, 
the VNGE, denoted by $H(G)$, is defined by replacing $\bPhi$ with $\bLNN=c \cdot \bL$ \citep{braunstein2006laplacian,passerini2008neumann,passerini2009quantifying}, where $c=1/\trace(\bL)$ is a trace normalization factor. It has been proved in \citep{passerini2008neumann} that for any $G \in \cG$,  $H(G) \leq \ln (n-1)$, where the equality holds when $G$ is a complete graph.
Note that since computing VNGE requires the entire eigenspectrum $\{\lambda_i\}_{i=1}^n$ of $\bLNN$, it incurs full eigenvalue decomposition on $\bLNN$ and has cubic complexity $O(n^3)$\footnote{$f(n)=O(h(n))$, $f(n)=o(h(n))$ and $f(n)=\Omega(h(n))$
	mean $\limsup_{n\ra \infty} |\frac{f(n)}{h(n)}|< \infty $, $\lim_{n\ra \infty} \frac{f(n)}{h(n)}=0 $, and  $\limsup_{n\ra \infty} |\frac{f(n)}{h(n)}|> 0 $, respectively. }\footnote{For computing all eigenvalues of large matrices, a viable solution is
	direct methods, possibly with parallel eigensolvers for acceleration. The  complexity for computing $\{\lambda_i\}_{i=1}^n$ of $\bLNN$ is typically $O(n^2 + \frac{4}{3}n^3)$ \citep{bai2000templates}.}
\citep{HornMatrixAnalysis}, making it computationally infeasible for large graphs.

In what follows, we propose two types of approximate VNGE ($\hH$ and $\tH$) for the exact VNGE $H$, where $\hH$ and $\tH$ possess linear computation complexity and satisfy $\tH \leq \hH \leq H$. Depending on the data format and problem setup, $\hH$ is designed for fast computation of $H$ for a single graph, and $\tH$ is designed for online computation of $H$ based on incremental graph changes. Furthermore,  we derive approximation error and prove asymptotic equivalence relative to $H$ under mild conditions on the eigenspectrum $\{\lambda_i\}_{i=1}^n$ of $\bLNN$. Our proofs are given in the supplementary material.

\subsection{Approximation of von Neumann Graph Entropy}

Recall that computing $H=-\sum_{i=1}^n \lambda_i \ln \lambda_i$ requires $O(n^3)$ computation complexity. For computation acceleration, we first reduce its computation complexity by using the quadratic approximation of the term $\lambda_i \ln \lambda_i$ in $H$ via Taylor series expansion, leading to the following lemma.
\begin{lem}[Quadratic approximation $Q$ of $H$] 
	\label{lemma_quad_approx}
	For any $G \in \cG$, the quadratic approximation $Q$ of the von Neumann graph entropy $H$ via Taylor series expansion is equivalent to 	$Q=1 - c^2 (\sum_{i \in \cV} s_i^2 + 2  \sum_{(i,j) \in \cE} w_{ij}^2)$, where $c=\frac{1}{S}$ and $S=
	\trace(\bL)=\sum_{i \in V} s_i=2\sum_{(i,j)\in \cE} w_{ij}$.
\end{lem}

It is clear from Lemma \ref{lemma_quad_approx} that $Q$ only depends on the edge weights in $G=(\cV,\cE,\bW)$, resulting in
linear computation complexity\footnote{The complexity becomes $O(n^2)$ when $m=O(n^2)$ (i.e., dense graphs). In sparse graphs $m$ could be $O(n)$.} $O(n+m)$, where  $|\cV|=n$ and $|\cE|=m$.
We note that higher-order (beyond quadratic) approximation of $H$ is plausible at the price of less computational efficiency and possibly excessive subgraph pattern searching. For example, the cubic approximation of $H$ involves the computation of $\trace(\bW^3)$, which relates to the sum of edge weights of every triangle in $G$.
To identify the approximation accuracy and equivalence of $Q$ with respect to $H$,
the following theorem shows the approximation bounds on $H$ in terms of $Q$ and the eigenspectrum $\{\lambda_{i}\}_{i=1}^n$ of $\bLNN$.

\begin{thm}[Approximation bounds on $H$]
	\label{thm_entropy_bound}
	For any $G \in \cG$, let $\lambda_{\max}$ and $\lambda_{\min}$ be the largest and smallest positive eigenvalue of $\bLNN$, respectively. If $\lambda_{\max}<1$, 
	then $ - Q \frac{ \ln \lambda_{\max}}{1-\lambda_{\min}} \leq H \leq -Q \frac{ \ln \lambda_{\min}}{1-\lambda_{\max}}$. The bounds become exact  and $H=\ln(n-1)$ when $G$ is a complete graph with identical edge weight.
\end{thm}
Note that Theorem \ref{thm_entropy_bound} excludes the extreme case when $\lambda_{\max}=1$, as the resulting VNGE is trivial ($H=0$). The condition $\lambda_{\max}<1$ holds for any graph $G \in \cG$ having a connected subgraph with at least 3 nodes.
In addition to the approximation bounds presented in Theorem \ref{thm_entropy_bound}, the corollary below further shows asymptotic equivalence between $Q$ and $\frac{H}{\ln n}$ under mild conditions on $\lambda_{\max}$ and $\lambda_{\min}$. 

\begin{cor}[Asymptotic equivalence of $Q$]
	\label{cor_asymptotic_Q}
	For any $G \in \cG$, let $n_+$ denote the number of positive eigenvalues of $\bLNN$.  If $n_+ =  \Omega(n)\footnotemark[1]$ and $\lambda_{\min} = \Omega(\lambda_{\max})$, then $\frac{H}{\ln n} - Q \ra 0 $ as $n \ra \infty$.
\end{cor}

Corollary \ref{cor_asymptotic_Q} suggests that the VNGE of large graphs with  balanced eigenspectrum (i.e.,  $\lambda_{\min}= \Omega(\lambda_{\max})  $) can be well approximated by $Q$ and a factor $\ln n$. The condition of  balanced eigenspectrum holds in regular and homogeneous random graphs \citep{passerini2008neumann,du2010note}. Furthermore, since $n_+$ equals to $n-g$, where $g$ is the number of connected components in $G$ \citep{Merris94}, the condition $n_+ = \Omega ( n)$ holds when $g=o(n)$\footnotemark[1].

\subsection{FINGER-$\hH$: Approximate von Neumann Graph Entropy $\hH$ Using $Q$ and $\lambda_{\max}$ }
\label{subsec_hH}
Based on the derived lower bound of $H$ in Theorem \ref{thm_entropy_bound}, we propose the first type of approximate VNGE $\hH$ using $Q$ and $\lambda_{\max}$ for any $G \in \cG$, which is defined as 
\begin{align}
\label{eqn_H_hat}
\hH(G)=-Q \ln \lambda_{\max}.
\end{align}
Comparing to the lower bound $-Q \frac{\ln \lambda_{\max}}{1-\lambda_{\min}}$ in  Theorem \ref{thm_entropy_bound}, $\hH$ is a looser lower bound on $H$ since $1-\lambda_{\min} < 1$. Here we use $1-\lambda_{\min} \approx 1$ when approximating $H$, since $\trace(\bLNN)=\sum_{i=1}^n \lambda_i=1$ and hence $\lambda_{\min}$ is negligible, especially for large graphs.

More importantly, since $\lambda_{\max}$ is the largest eigenvalue of $\bLNN$ and by definition  $\bLNN$ has $n+m$ nonzero entries, the computation of $\lambda_{\max}$ only requires $O(m+n)$ operations via power iteration methods \citep{HornMatrixAnalysis,wu2017primme_svds,liao2019lanczosnet}, leading to the same complexity as $Q$. Consequently, by only acquiring  $\lambda_{\max}$  instead of the entire eigenspectrum $\{\lambda_i\}_{i=1}^n$,
the computation of $\hH$ has linear complexity  $O(m+n)$, resulting in significant computation reduction when compared with the exact VNGE $H$, which requires cubic complexity\footnotemark[2] $O(n^3)$. In addition to computational efficiency, the following corollary shows that the approximation error of $\hH$, defined as $H-\hH$, decays at the rate of $\ln n$ under the same conditions as in Corollary \ref{cor_asymptotic_Q}. We note that the $o(\ln n)$ approximation error rate is nontrivial since $H \leq \ln(n-1)$ for any $G \in \cG$ \citep{passerini2008neumann,du2010note}.

\begin{cor}[$o(\ln n)$ approximation error of $\hH$]
	\label{cor_approx_error_hH}
	For any $G \in \cG$, if $n_+ =\Omega(n)$ and $\lambda_{\min} = \Omega(\lambda_{\max}) $, then the scaled approximation error (SAE)
	$\frac{H - \hH }{\ln n} \ra 0$ as $n \ra \infty$, implying $H - \hH=o(\ln n)$. 
\end{cor}

\subsection{FINGER-$\tH$: Approximate von Neumann Graph Entropy $\tH$ Using $Q$ and $s_{\max}$}
\label{subsec_tH}
The proxy $\hH$ in Section \ref{subsec_hH} enables fast computation of VNGE for a single graph.  As the exact online update of the eigenvalue $\lambda_{\max}$ in $\hH$ based on incremental graph changes is challenging, 
we propose the second type of approximate VNGE $\tH$ using $Q$ and the largest nodal strength $s_{\max}=\max_{i \in \cV} s_i$ in a graph, which allows simple incremental update of $\tH$ based on graph changes but at the price of larger approximation error than that of $\hH$. The approximate VNGE $\tH$ is defined as
\begin{align}
\label{eqn_tH}
\tH(G)= -Q \ln ( 2 c \cdot  s_{\max} ),
\end{align}
where $c$ is the trace normalization constant. 
Using the definition $\bLNN=c \cdot \bL$ and the upper bound on the largest eigenvalue of $\bL$ in \citep{anderson1985eigenvalues}, we obtain $\tH \leq \hH \leq H$ since $\lambda_{\max} \leq 2 c \cdot s_{\max}$, implying $\tH$ is a looser lower bound on $H$ when compared with $\hH$. Nonetheless, the following corollary shows the approximation error of $\tH$ also decays at the same rate   $o(\ln n)$ as $\hH$.

\begin{cor}[$o(\ln n)$ approximation error of $\tH$]
	\label{cor_approx_error_tH}	
	For any $G \in \cG$, if $n_+ = \Omega(n)$ and $\lambda_{\min} = \Omega(\lambda_{\max}) $,
	then the scaled approximation error (SAE)	$\frac{H - \tH }{\ln n} \ra 0$ as $n \ra \infty$, implying $H - \tH=o(\ln n)$. 
\end{cor}

To enable incremental computation of VNGE using $\tH$, let $G=(\cV,\cE,\bW)$ and $\pG=(\cpV,\cpE,\bpW)$ be any two graphs from a graph sequence. Without loss of generality we assume $G$ and $\pG$ have a common node set $\cV_c$ with $|\cV_c|=n$ nodes\footnote{If $G$ and $\pG$ have different nodes, the set $\cV_c$ can be constructed by the set union  $\cV_c= \cV \cup \cpV$.}. In particular, the graph $\Delta G=(\Delta \cV, \Delta \cE,\Delta \bW)$ with $|\Delta \cV|=\Delta n$ and $|\Delta \cE|=\Delta m$ is introduced to represent the changes made from converting $G$ to $\pG$, denoted by $\pG=G \oplus \Delta G$\footnote{The notation $\oplus$ denotes set additions $\cpV  = \cV \biguplus \Delta V$, $\cpE  = \cE \biguplus  \Delta \cE$ and matrix addition  $\bpW=\bW + \Delta \bW$.}. The terms $\{\Delta s_i\}_{i \in \Delta \cV}$ and  $\{\Delta w_{ij}\}_{(i,j) \in \Delta \cE}$ denote the nodal strengths and edge weights of $\Delta G$, respectively, and $\Delta S=\sum_{i \in \Delta \cV} \Delta s_i$. Let $\pQ$ be the quadratic approximation of $H(\pG)$. The theorem below shows that $\pQ$ can be efficiently updated based on  $Q$ of $H(G)$, the values of $s_{\max}$ and $c$ from $G$, and $\Delta G$, yielding competent complexity $O(\Delta n + \Delta m)$.

\begin{thm}[Incremental update of $\pQ$]
	\label{thm_incremental}
	For any $G, \pG \in \cG$ such that $\pG=G \oplus \Delta G$, given $Q$, $G$ and $\Delta G$, the term $\pQ$ can be efficiently updated by incremental graph changes as
	$\pQ=\frac{Q-1}{(1+ c \Delta S )^2}   - \lb \frac{c}{1+c \Delta S} \rb^2 \Delta Q   +1$,
	where  $\Delta Q = 2 \sum_{i \in \Delta \cV } s_i \Delta s_i + \sum_{i \in \Delta \cV }  \Delta s_i^2 + 4 \sum_{(i,j) \in \Delta \cE} w_{ij} \Delta w_{ij} + 2\sum_{(i,j) \in \Delta \cE} \Delta w_{ij}^2 $, and $\Delta c= \frac{- c^2 \Delta S}{1+ c \Delta S}$.
\end{thm}

Furthermore, by the definition of $\tH$ in (\ref{eqn_tH}), $\tH(G \oplus \Delta G)$ can be efficiently updated by
\begin{align}
\label{eqn_tH_inc}
\tH(G \oplus \Delta G)=- \pQ \ln [ 2 (c+\Delta c) (s_{\max}+\Delta s_{\max}) ]
\end{align} 
given $Q$, $s_{\max}$ and $c$ from $G$, and graph changes $\Delta G$,
where $\Delta c$ is defined in Theorem \ref{thm_incremental}, and $\Delta s_{\max}$ is the maximum value of $0$ and $\max_{i \in \Delta \cV} (s_i+ \Delta s_i)-s_{\max}$.
The computation complexity of   $\tH(G \oplus \Delta G)$ is $O(\Delta n + \Delta m)$ since
the incremental update formula of  $\pQ$ in Theorem \ref{thm_incremental} and the computation of $\Delta s_{\max}$
only take  $O(\Delta n + \Delta m)$ operations.

\begin{algorithm}[t]
	\caption{FINGER-JSdist (Fast)}
	\label{algo_JSfast}
	\begin{algorithmic}
		   \STATE \textbf{Input:} Two graphs $G$ and $\pG$ from a graph sequence
		   \STATE \textbf{Output:} \textsf{JSdist}($G, \pG$)
		   \STATE 1. Obtain $\overline{G}=\frac{G \oplus \pG}{2}$ and compute $\hH(G)$, $\hH(\pG)$, and $\hH(\overline{G})$ via FINGER-$\hH$ from (\ref{eqn_H_hat})
		   \STATE 2.   $\textsf{JSdist}(G,\pG)=\lb \hH(\overline{G})- \frac{1}{2} [\hH(G) + \hH(\pG)] \rb^{1/2}$
	\end{algorithmic}
\end{algorithm} 

\begin{algorithm}[t]
	\caption{FINGER-JSdist (Incremental)}
	\label{algo_JSinc}
	\begin{algorithmic}
		   \STATE \textbf{Input:} A graph $G$,  graph changes $\Delta G$, and $\tH(G)$ 
		   \STATE \textbf{Output:} \textsf{JSdist}($G,G \oplus \Delta G$)
		\STATE 1. Compute $\tH(G \oplus \frac{\Delta G}{2})$ and  $\tH(G \oplus \Delta G)$ via FINGER-$\tH$ from (\ref{eqn_tH_inc}) and Theorem \ref{thm_incremental}
		   \STATE 2.  $\textsf{JSdist}(G,G \oplus \Delta G)=$
		   \STATE $\lb \tH(G \oplus \frac{\Delta G}{2})- \frac{1}{2} [\tH(G) + \tH(G \oplus \Delta G)] \rb^{1/2}$
	\end{algorithmic}
\end{algorithm}

\subsection{Fast and Incremental Algorithms for Jensen-Shannon Distance between Graphs}
\label{subsec_FINGERJS}
As summarized in Algorithms  \ref{algo_JSfast} and  \ref{algo_JSinc},
one major utility of VNGE\footnote{Codes: \url{https://github.com/pinyuchen/FINGER}} is the computation of Jensen-Shannon distance (JSdist) between any two graphs from a graph sequence. Consider two graphs $G=(\cV_c,\cE,\bW) \in \cG$ and $\pG=(\cV_c,\cpE,\bpW) \in \cG$,  and let $\oG=(\cV_c,\coE,\boW)=\frac{G \oplus \pG}{2}$ denote their averaged graph such that $\boW=\frac{\bW+\bpW}{2}$. Then the Jensen-Shannon divergence between $G$ and $\pG$ can be computed by $\textsf{JSdiv}(G,\pG)=H(\oG)-\frac{1}{2}[H(G)+H(\pG)]$ \citep{de2015structural}. Furthermore, the Jensen-Shannon distance between $G$ and $\pG$ is defined as $\textsf{JSdist}(G,\pG)=\sqrt{\textsf{JSdiv}(G,\pG)}$, which has been proved to be a valid distance metric in \citep{endres2003new,briet2009properties}.
The exact computation of $\textsf{JSdist}$ requires $O(n^3)$ computation complexity by the definition of $H$, where $|\cV_c|=n$, which is computationally cumbersome for large graphs. To overcome its computational inefficiency, we apply the developed FINGER-$\hH$ and  FINGER-$\tH$ to the computation of  $\textsf{JSdist}$.
If each graph $G_t$ in a graph sequence $\{G_t\}_{t=1}^T$ is given, then FINGER-JSdist (Fast) allows fast computation of $\textsf{JSdist}$ and features linear computation complexity inherited from $\hH$. If a graph sequence is presented by sequential graph changes $\{\Delta G_t\}_{t=1}^{T-1}$ such that $G_{t+1} = G_{t} \oplus \Delta G_t$,  then FINGER-JSdist (Incremental) allows online computation of $\textsf{JSdist}$ relative to the incremental graph changes.
Their superior performance will be discussed in Section \ref{sec_app}.

\begin{figure*}[t]
	\centering
	\begin{subfigure}[b]{0.32\linewidth}
		\includegraphics[width=\textwidth]{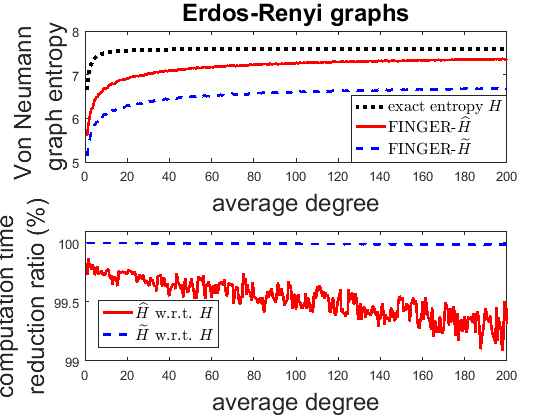}
		\vspace{-5mm}
		\caption{ER model}
	\end{subfigure}%
	\centering
	\begin{subfigure}[b]{0.32\linewidth}
		\includegraphics[width=\textwidth]{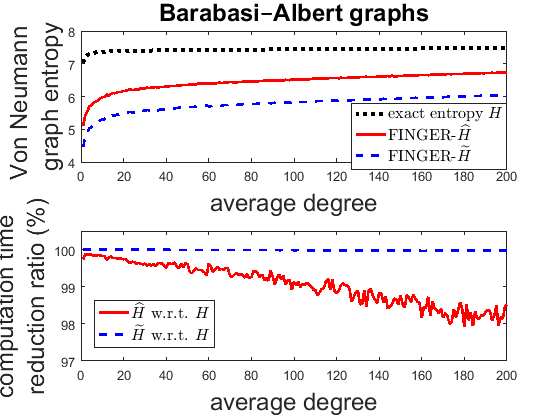}
		\vspace{-5mm}
		\caption{BA model}
	\end{subfigure}
	\centering
	\begin{subfigure}[b]{0.32\linewidth}
		\includegraphics[width=\textwidth]{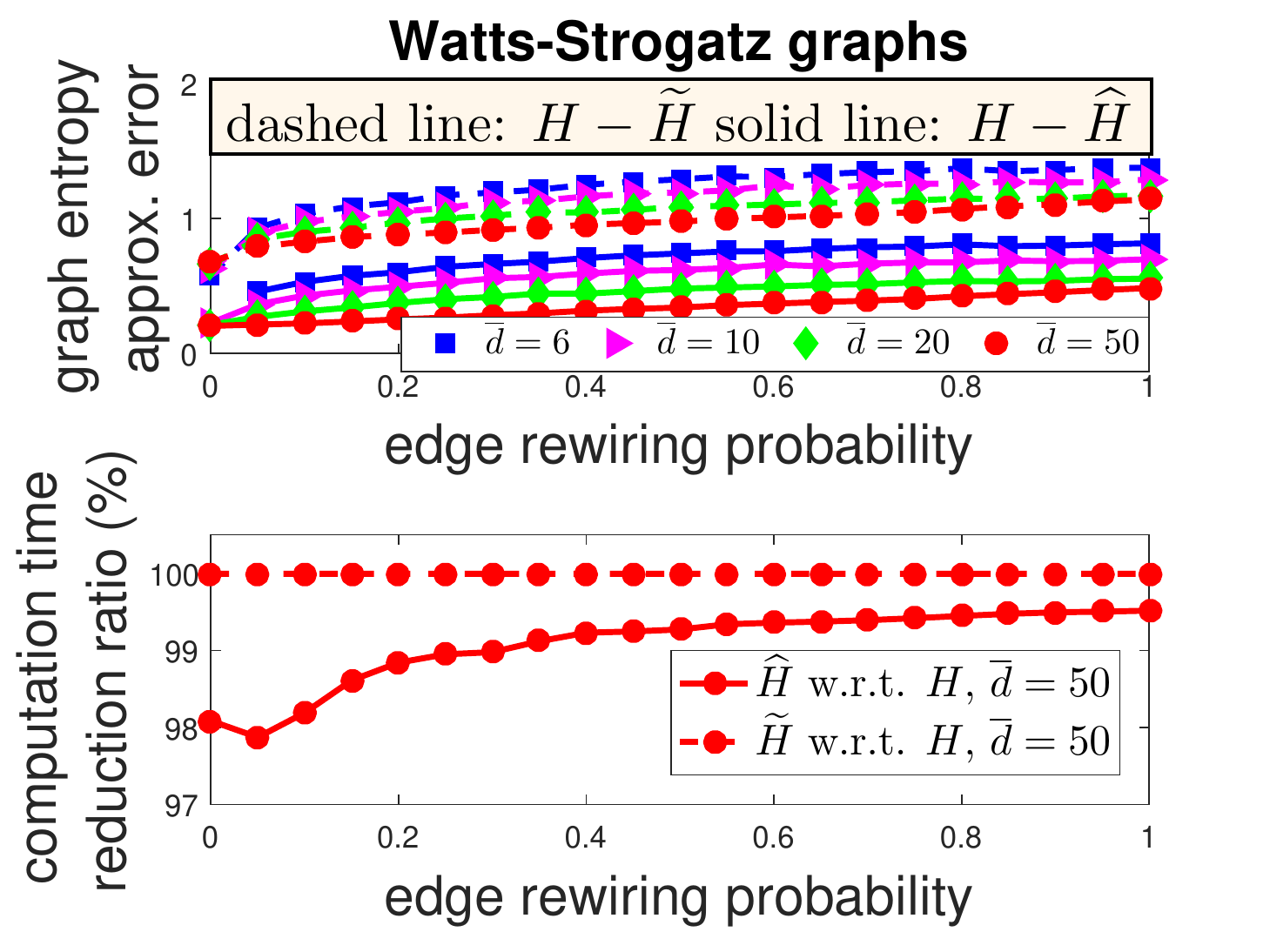}
		\vspace{-5mm}
		\caption{WS model}
	\end{subfigure}	
	\vspace{-4mm}	
	\caption{Performance evaluation of von Neumann graph entropy approximation in different random graph models with $n=2000$ nodes under varying average degree $\od$ and edge rewiring probability $\pWS$. The approximation error of FINGER decays as $\od$ increases or $\pWS$ decreases. FINGER achieves nearly 100\% speed-up relative to the  exact entropy computation.}
	\label{Fig_FINGER_deg}
	\vspace{-2mm}
\end{figure*}

\begin{figure*}[t]
	\centering
	\begin{subfigure}[b]{0.32\linewidth}
		\includegraphics[width=\textwidth]{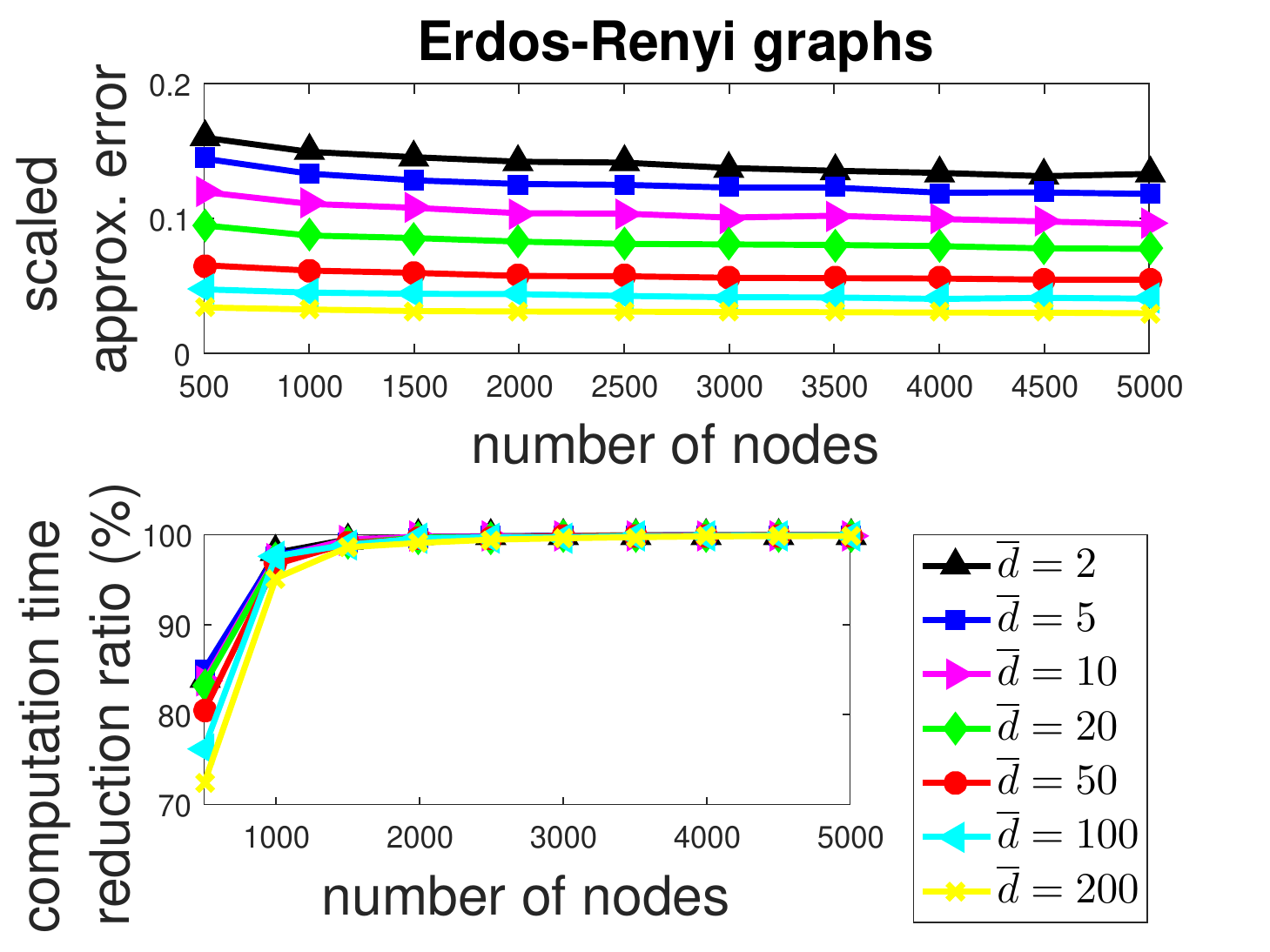}
		\vspace{-6mm}
		\caption{ER model}
		\label{Fig_ER_FINGER}
	\end{subfigure}%
	\centering
	\begin{subfigure}[b]{0.32\linewidth}
		\includegraphics[width=\textwidth]{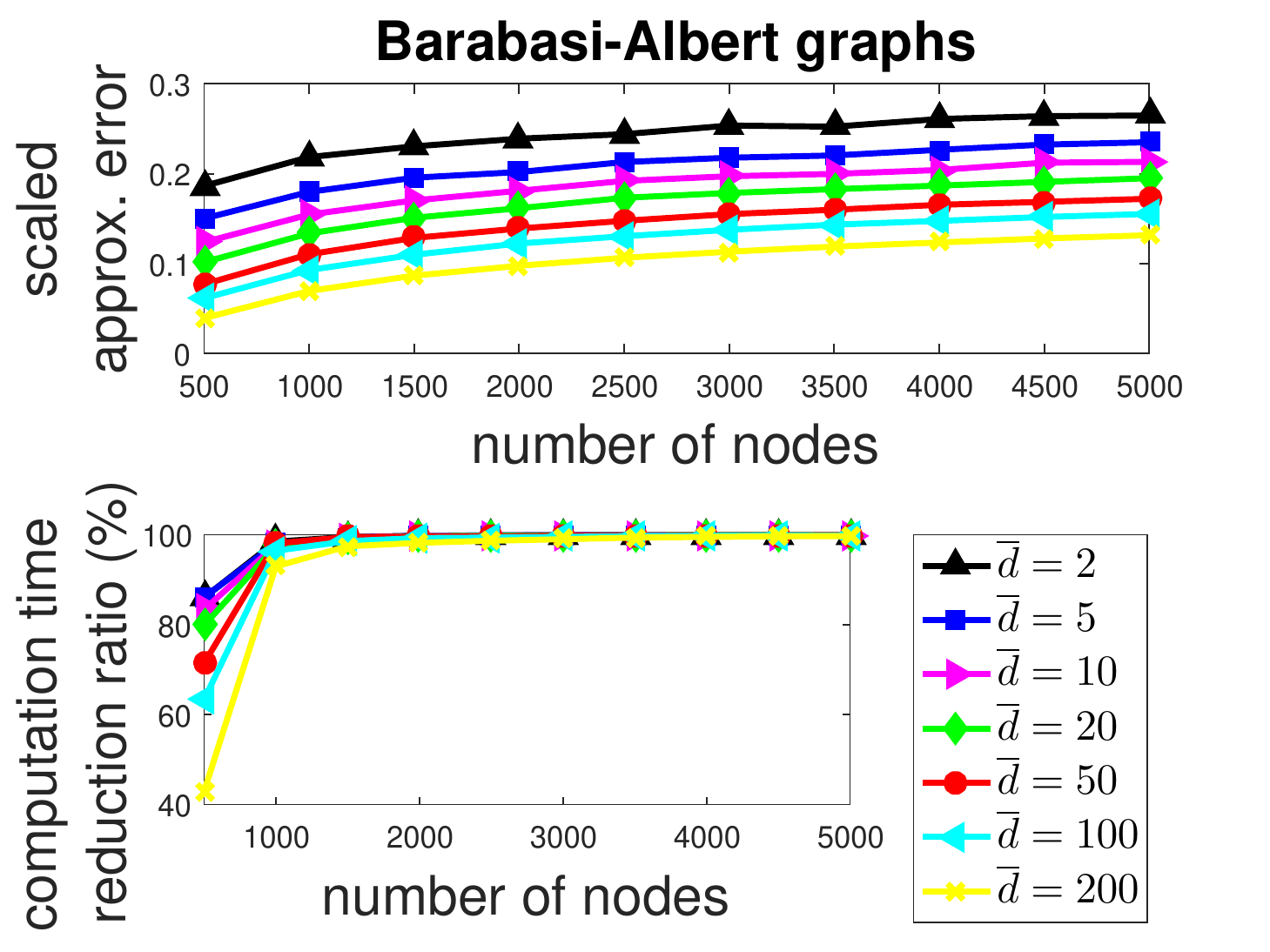}
		\vspace{-6mm}
		\caption{BA model}
	\end{subfigure}
	\centering
	\begin{subfigure}[b]{0.32\linewidth}
		\includegraphics[width=\textwidth]{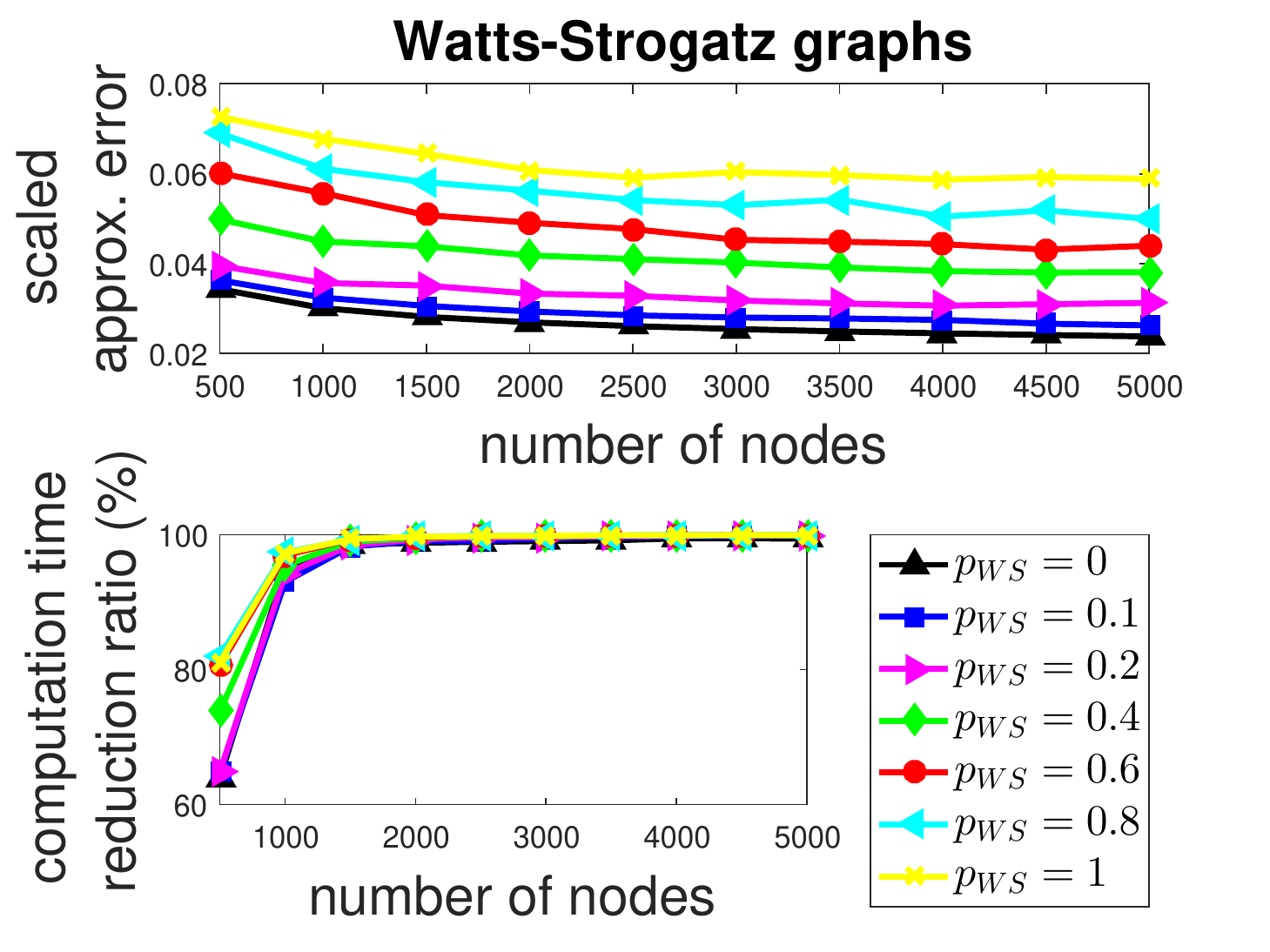}
		\vspace{-6mm}
		\caption{WS model ($\od=50$)}
	\end{subfigure}	
	\vspace{-4mm}				
	\caption{Scaled approximation error (SAE) and computation time reduction ratio (CTRR) of $\hH$ via FINGER for different random graph models and varying number of nodes $n$. The SAE of ER and WS graphs validates the $o(\ln n)$ approximation error analysis in  Corollaries \ref{cor_approx_error_hH} and \ref{cor_approx_error_tH}. The CTRR attains nearly 100\% speed-up relative to $H$ for  moderate-size graphs ($n \geq 2000$).}
	\label{Fig_FINGER_node}
\end{figure*}

\section{Experiments}
\label{sec_experiemnts}
In this section we conducted intensive experiments on the VNGE of three kinds of synthetic random graphs to study the effects of graph size, average degree, and graph regularity on the  approximation error of FINGER and its computational efficiency. The three random graph models are: (i) Erdos-Renyi (ER) model \citep{erdos1959random} -- every node pair is connected independently with probability $\pER$; (ii) Barabasi-Albert (BA) model \citep{Barabasi99} -- the degree distribution follows a power-law distribution; and (iii) Watts-Strogatz (WS) model \citep{Watts98_simple} -- an initially regular ring network with independent edge rewiring probability $\pWS$ for simulating small-world networks. The parameter $\pWS$ controls the regularity of graph connectivity, and smaller $\pWS$ gives more regular graphs. 

Since $\tH \leq \hH \leq H$, the approximation error (AE) is defined as $H-\hH$ and $H-\tH$, respectively. The scaled approximation error (SAE) is defined as $\frac{\textnormal{AE}}{\ln n}$, which is a proper scaling according to our error analysis in Section \ref{Sec_FINGER}, and it also makes a fair comparison of graphs with different number of nodes.   The computation time reduction ratio (CTRR) is defined as $\frac{\textnormal{Time}(H)-\textnormal{Time}(X)}{\textnormal{Time}(H)}$, where $X \in \{ \hH, \tH\}$ and $\textnormal{Time}(Y)$ denotes the computation time for $Y \in \{H,\hH,\tH\}$. All experiments (including Section \ref{sec_app}) were conducted by Matlab R2016 on a 16-core machine with 128 GB RAM. The results in this section are averaged over 10 random trials.
Additional results are reported in the supplementary material.

\textbf{The effect of average degree  $\od$ and graph regularity parameter $\pWS$.}
Figures \ref{Fig_FINGER_deg} (a) and \ref{Fig_FINGER_deg} (b) display the exact and the two approximate VNGE of ER and BA graphs and the corresponding CTRR under varying $\od$. When fixing the number of nodes $n$, both $\hH$ and $\tH$ better match $H$ as  $\od$ increases, suggesting their
AE decays with $\od$. Comparing their CTRR, the computation of $\hH$ and $\tH$ enjoys at least $97\%$ speed-up relative to $H$. The drastic reduction in computation time can be explained by the efficient linear complexity of FINGER, as opposed to the high complexity in computing the entire eigenspectrum for calculating $H$. The CTRR of $\hH$ slightly decays  with $\od$ due to the growing number of nonzero entries (edges) in $\bLNN$, resulting in increasing operations for computing $\lambda_{\max}$. Although the AE of $\hH$ is always smaller than that of $\tH$ due to the fact that $\tH \leq \hH \leq H$, the CTRR of $\tH$ has nearly 100\% speed-up relative to $H$ by simply requiring the information of $s_{\max}$ instead of $\lambda_{\max}$ from a graph.

Figure \ref{Fig_FINGER_deg} (c) displays the AE and CTRR of $\hH$ and $\tH$ under varying edge rewiring probability $\pWS$ and different average degree $\od \in \{6,10,20,50\}$ of WS model. Similar to ER and BA graphs, when fixing $n$ and $\pWS$, the AE of $\hH$ and  $\tH$ decays as $\od$ increases. When $n$ and $\od$ are fixed, smaller $\pWS$ yields less AE for both $\hH$ and $\tH$, suggesting that FINGER attains better approximation when graphs are more regular. Since the curves of CTRR for different $\od$ in WS model have similar behavior, here we only report the results when $\od=50$. Consistent with the observations in ER and BA graphs, in WS graphs the CTRR of $\hH$ and $\tH$  achieves nearly 100\% improvement relative to $H$, and $\tH$ attains slightly better CTRR than $\hH$ at the price of larger AE. 

\textbf{The effect of graph size $n$.} 
Figure \ref{Fig_FINGER_node} displays the SAE of FINGER under the three random graph models when varying the number of nodes $n$. Since the results of $\hH$ and $\tH$ are similar, 
we show the SAE of $\hH$ in Figure \ref{Fig_FINGER_node} and report the SAE of $\tH$ in the supplementary material.
By the fact that ER and WS graphs have balanced eigenspectrum  \citep{Mieghem10}, for ER and WS models the SAE of both $\hH$ and $\tH$ decays as $n$ increases, which verifies the $o(\ln n)$ approximation error as stated in Corollaries \ref{cor_approx_error_hH} and \ref{cor_approx_error_tH}.  On the other hand, the SAE of BA graphs is observed to grow logarithmically in $n$ due to the existence of extreme eigenvalues (imbalanced eigenspectrum) \citep{Mieghem10,goh2001spectra}. Similar to the observations from fixed-size graphs, for a fixed $n$ the SAE decays with $\od$ and graph regularity  in all cases. In addition, the CTRR attains nearly 100\% speed-up relative to $H$ for  moderate-size graphs ($n \geq 2000$). 

\section{Applications}
\label{sec_app}
Here we apply FINGER to the computation of Jensen-Shannon (JS) distance between graphs  (Section \ref{subsec_FINGERJS}) in two  applications and one synthesized dataset and demonstrate its outstanding performance over seven baseline and state-of-the-art methods in terms of efficiency and effectiveness.

\begin{table*}[t]
	\centering
	\caption{Summary of four evolving Wikipedia hyperlink networks.}
	\label{table_wiki}
	\begin{adjustbox}{max width=0.85\textwidth}		
		\begin{tabular}{llll}		
			\toprule
			Datasets (graph sequence) & maximum \# of nodes & maximum \# of edges & \# of graphs \\ 
			\midrule
			Wikipedia - simple English (sEN)       & 100,312 (0.1 M)     & 746,086  (0.7 M)      & 122          \\
			Wikipedia - English  (EN)        & 1,870,709  (1.8 M)       & 39,953,145 (39 M)     & 75  \\		
			Wikipedia - French (FR)       & 2,212,682  (2.2 M)     & 24,440,537 (24 M)    & 121          \\
			Wikipedia - German (GE)       & 2,166,669 (2.1 M)      & 31,105,755 (31 M)    & 127          \\
			\bottomrule
		\end{tabular}
	\end{adjustbox}
	\vspace{-2mm}
\end{table*}

\begin{table*}[t]
	\centering	
	\caption{Computation time (seconds) and Pearson correlation coefficient (PCC) between the anomaly proxy and different methods. FINGER attains the best PCC and time efficiency. The  Spearman's rank correlation analysis is given in Table \ref{table_SRCC} of the supplement. 
	}
	\label{table_wiki_PCC}
	\begin{adjustbox}{max width=0.94\textwidth}		
		\begin{tabular}{lllllllllll}
			\hline
			\multicolumn{2}{c}{Datasets}                                                 & \begin{tabular}[c]{@{}l@{}}FINGER\\ -JS (Fast)\end{tabular} & \begin{tabular}[c]{@{}l@{}}FINGER\\ -JS (Inc.)\end{tabular} & DeltaCon & RMD     & \begin{tabular}[c]{@{}l@{}}$\lambda$ dist. \\ (Adj.)\end{tabular} & \begin{tabular}[c]{@{}l@{}}$\lambda$ dist. \\ (Lap.)\end{tabular} & GED     & \begin{tabular}[c]{@{}l@{}}VNGE\\ -NL\end{tabular} & \begin{tabular}[c]{@{}l@{}}VNGE\\ -GL\end{tabular} \\ \hline
			\multirow{2}{*}{\begin{tabular}[c]{@{}l@{}}Wiki\\ (sEN)\end{tabular}} & PCC  & \textbf{0.5593}                                             & 0.3382                                                      & 0.1596   & 0.1718  & 0.1871                                                            & -0.0095                                                           & -0.2036 & 0.2065                                             & 0.2462                                             \\
			& time & 26.065                                                      & \textbf{0.7438}                                             & 44.952   & 44.952  & 150.16                                                            & 99.905                                                            & 1.666   & 13.574                                             & 30.483                                             \\ \hline
			\multirow{2}{*}{\begin{tabular}[c]{@{}l@{}}Wiki\\ (EN)\end{tabular}}  & PCC  & \textbf{0.9029}                                             & 0.5583                                                      & -0.2411  & -0.1167 & -0.0175                                                           & -0.1759                                                           & -0.3429 & -0.0442                                            & 0.1519                                             \\
			& time & 603.98                                                      & \textbf{13.975}                                             & 1846.1   & 1846.1  & 4417.7                                                            & 2898.3                                                            & 47.299  & 335.66                                             & 858.22                                             \\ \hline
			\multirow{2}{*}{\begin{tabular}[c]{@{}l@{}}Wiki\\ (FR)\end{tabular}}  & PCC  & \textbf{0.8183}                                             & 0.592                                                       & -0.1503  & -0.1203 & 0.0133                                                            & -0.1877                                                           & -0.4915 & 0.0552                                             & 0.2349                                             \\
			& time & 1038.6                                                      & \textbf{23.667}                                             & 2804.5   & 2804.5  & 6664.5                                                            & 4411.4                                                            & 83.398  & 474.42                                             & 1129.1                                             \\ \hline
			\multirow{2}{*}{\begin{tabular}[c]{@{}l@{}}Wiki\\ (GE)\end{tabular}}  & PCC  & \textbf{0.6764}                                             & 0.4619                                                      & -0.2035  & -0.1542 & 0.0182                                                            & -0.3814                                                           & -0.4677 & 0.2194                                             & 0.2679                                             \\
			& time & 1457.3                                                      & \textbf{32.647}                                             & 4184.1   & 4184.1  & 9462.5                                                            & 6013.7                                                            & 115.923 & 716.31                                             & 1674.6                                             \\ \hline
		\end{tabular}
	\end{adjustbox}
\end{table*}


\textbf{Anomaly detection in evolving Wikipedia hyperlink networks.}
Wikipedia is an online encyclopedia that allows editing and referencing between articles. By viewing an article as a node and a hyperlink as an edge, the evolution of Wikipedia forms a graph sequence $\{G_t\}_{t=1}^T$ over time. Table \ref{table_wiki} summarizes four evolving Wikipedia networks of different language settings collected in \citep{mislove2009online,preusse2013structural}, where each graph $G_t=(\cV_t,\cE_t,\bW_t)$ corresponds to a monthly snapshot of a hyperlink network. These datasets are presented in terms of addition and deletion of nodes or edges with timestamps (i.e., continuous graph changes $\{\Delta G_t\}_{t=1}^{T-1}$), which directly applies to incremental JS distance computation via FINGER (Algorithm \ref{algo_JSinc}). Fast JS distance computation via FINGER (Algorithm \ref{algo_JSfast}) can also be applied by
computing $G_{t+1}=G_t \oplus \Delta G_t$ to obtain $\{G_t\}_{t=1}^T$. The task of anomaly detection is to identify noticeable changes (relative to the bulk network) in the consecutive monthly snapshots of these massive Wikipedia hyperlink networks.

\textbf{Bifurcation detection in dynamic genomic networks.}
The genome-wide chromosome conformation capture (Hi-C) contact maps \citep{beloqui2009reactome} for studying cell reprogramming from human fibroblasts to skeletal muscle can be viewed as a graph sequence consisting of 12 sampled spatial measurements, in which the cell reprogramming undergoes a space-time bifurcation at the 6th measurement as verified in \citep{liu2018genome}. The task is to identify this bifurcation instance based on the dynamic  Hi-C contact maps.
Additional descriptions of this dataset are given in the supplementary material.

\textbf{\textbf{Evaluation.}} 
We note that there are two major differences between these two applications: (i) unweighted v.s.
weighted graphs and (ii) with v.s. without ground truth. \\
In the Wikipedia case (unweighted graphs), our main goal is to use these large datasets to demonstrate the efficient computation  of JS distance via FINGER owing to its linear complexity. Additionally, since there are no labels for verifying the detected changes, we conduct an ex post facto correlation analysis using an explicit and explainable anomaly metric -- the vertex/edge overlapping (VEO) score \citep{papadimitriou2010web}. VEO is a properly normalized metric reflecting topological differences between two unweighted graphs, defined as $1-\frac{2(|\cV_t \cap \cV_{t+1}|+|\cE_t \cap \cE_{t+1}|)}{|\cV_t|+|\cV_{t+1}|+|\cE_{t}|+|\cE_{t+1}|}$, which is between $[0,1]$ and  relates to the Sorensen–Dice coefficient \citep{dice1945measures,sorensen1948method} for comparing the similarity of two samples. In the Wikipedia experiments, a high VEO score directly pinpoints the month when articles are edited by a relatively significant amount. Consequently, VEO can be used as an anomaly proxy for ex post facto analysis in our setting.
\\
In the genome case (weighted graphs), the ground-truth bifurcation instance was verified. Moreover, unlike the Wikipedia case, the genome dataset contains nonnegative edge weights indicating cell interaction strengths. Therefore, in this case VEO is not an appropriate anomaly proxy because by definition it is insensitive to edge weight changes.


\textbf{Comparative methods.} We compare the proposed method with the following baseline methods: 

$\bullet$ DeltaCon \citep{koutra2016d}: DeltaCon  uses the idea of fast belief propagation to compute graph similarity and outputs a similarity score $\textnormal{Sim}_{\textnormal{DC}} \in [0,1]$. We use $1-\textnormal{Sim}_{\textnormal{DC}}$ as the anomaly score.\\ 
$\bullet$ RMD \citep{koutra2016d}: RMD is the Matusita distance deduced from DeltaCon, which is defined as $\frac{1}{\textnormal{Sim}_{\textnormal{DC}}} -1$.\\
$\bullet$ $\lambda$ distance \citep{bunke2007graph,wilson2008study}: 
The Euclidean distance between two sets of top $k$ eigenvalues of a matrix. Here we use the weight matrix $\bW$ (Adj.) and the graph Laplacian matrix $\bL$ (Lap.), and set $k=6$.
\\
$\bullet$ GED \citep{bunke2007graph}: graph edit distance (GED) for undirected unweighted graphs is the number of operations (node/edge additions and removals) required to convert a graph $G_t$ to another graph $G_{t+1}$.
\\
$\bullet$  VNGE-NL \citep{han2012graph} / VNGE-GL \citep{ye2014approximate}: Two VNGE heuristics using the normalized/generalized graph Laplacian matrix. Unlike FINGER, they lack approximation error guarantee.


\textbf{Wikipedia results.} We compute the dissimilarity metrics of each method and compare them with the anomaly proxy in terms of the Pearson correlation coefficient (PCC). A higher PCC suggests a better match to the anomaly proxy for detecting abnormal monthly edit changes relative to the bulk network. The PCC and computation time of each method are reported in Table \ref{table_wiki_PCC}.
For illustration, the dissimilarity metrics of Wikipedia-EN are shown in Figure \ref{Fig_app_wiki}. 
The plots of the other Wikipedia networks are given in the supplementary material. 
The statistics of the anomaly proxy meet the intuition that in the earlier stage the monthly evolution of Wikipedia is more drastic, and in the later stage it becomes stable (i.e., less anomalous) since the changes are subtle relative to the entire network.  In Table \ref{table_wiki_PCC}, FINGER-JSdist (Fast) attains the best PCC (0.9029) and competitive computation time. This suggests that the computation of JS distance can be made efficient by FINGER, and its ex post facto analysis is highly correlated with the anomaly proxy. For example, in Figure \ref{Fig_app_wiki} their top 10 flagged anomalies have 9 months in common. On the other hand, the other dissimilarity metrics are either implicitly defined, unnormalized or lacking approximation guarantees, making the detected anomalies less explainable.
FINGER-JSdist (Incremental) has the least computation time by leveraging online computation, and it achieves the second best PCC due to looser approximation error of $\tH$ than $\hH$. 
Nonetheless, FINGER-JSdist (Incremental)  is roughly 3 times faster than GED, 20 times faster than VNGE-GL,
50 times faster than FINGER-JSdist (Fast), 100 times faster than DeltaCon, RMD and VNGE-NL, and 200-300 times faster than $\lambda$ distance. In addition to PCC, we also report the  rank correlation coefficients in the supplementary material to show the high correlation between FINGER and the anomaly proxy.

\begin{figure}[t]
		\vspace{-2mm}
	\centering
	\includegraphics[width=0.48\textwidth]{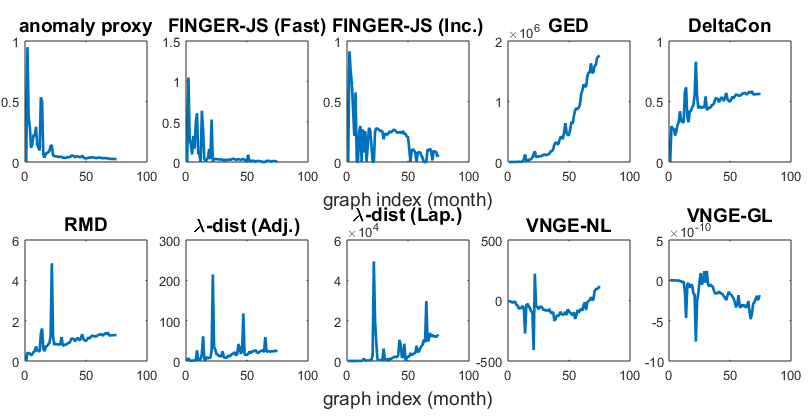}
	\vspace{-8mm}	
	\caption{ Dissimilarity (anomaly) metrics of consecutive monthly Wikipedia-English hyperlink networks. The ex post facto analysis shows FINGER-JSdist (Fast) is highly correlated with the anomaly proxy (0.9029 PCC in Table \ref{table_wiki_PCC} and 0.7973 SRCC in Table \ref{table_SRCC}).  FINGER-JSdist (Incremental) has efficient computation time and attains the second best PCC and SRCC among all methods.
	}
	\label{Fig_app_wiki}
	\vspace{-4mm}
\end{figure}

\begin{figure}[t]
	\centering
	\includegraphics[width=0.48\textwidth]{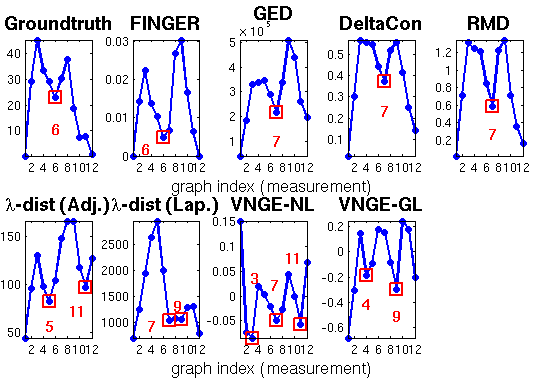}
	\vspace{-8mm}	
	\caption{Bifurcation detection of cell reprogramming in dynamic genomic networks via the temporal difference score (TDS) of different methods (y-axis). The red squares indicate the detected bifurcation points.  Among all the compared methods, FINGER-JSdist (Algorithm \ref{algo_JSfast}) is the only method that correctly detects the ground-truth bifurcation point (index 6), and its TDS resembles the shape of the ground-truth statistic.}
	\label{Fig_app_cell}
	\vspace{-2mm}
\end{figure}

\begin{table*}[!t]
	\centering
	\caption{Detection rate on synthesized anomalous events in the dynamic communication networks.}
	\label{table_oregon}
	\begin{adjustbox}{max width=0.9\textwidth}		
		\begin{tabular}{l|lllllllll}
			\hline
			\multicolumn{1}{c|}{DoS attack ($X\%$)} & \begin{tabular}[c]{@{}l@{}}FINGER\\ -JS (Fast)\end{tabular} & \begin{tabular}[c]{@{}l@{}}FINGER\\ -JS (Inc.)\end{tabular} & DeltaCon      & RMD           & \begin{tabular}[c]{@{}l@{}}$\lambda$ dist. \\ (Adj.)\end{tabular} & \begin{tabular}[c]{@{}l@{}}$\lambda$ dist. \\ (Lap.)\end{tabular} & GED  & \begin{tabular}[c]{@{}l@{}}VNGE\\ -NL\end{tabular} & \begin{tabular}[c]{@{}l@{}}VNGE\\ -GL\end{tabular} \\ \hline
			1 \%                                    & \textbf{24 \%}                                              & 10\%                                                        & 14\%          & 14\%          & 10\%                                                              & \textbf{24\%}                                                     & 14\% & 22\%                                               & 22\%                                               \\ \hline
			3 \%                                    & \textbf{75\%}                                               & 62\%                                                        & 58\%          & 58\%          & 12\%                                                              & 23\%                                                              & 36\% & 39\%                                               & 39\%                                               \\ \hline
			5 \%                                    & \textbf{90\%}                                               & 77\%                                                        & \textbf{90\%} & \textbf{90\%} & 12\%                                                              & 28\%                                                              & 41\% & 67\%                                               & 67\%                                               \\ \hline
			10 \%                                   & \textbf{91\%}                                               & \textbf{91\%}                                               & \textbf{91\%} & \textbf{91\%} & \textbf{91\%}                                                     & \textbf{91\%}                                                     & 81\% & \textbf{91\%}                                      & \textbf{91\%}                                      \\ \hline
		\end{tabular}
	\end{adjustbox}
		\vspace{-4mm}
\end{table*}

As discussed in the ``Evaluation'' paragraph, the main purpose of the Wikipedia experiments (without ground truths) is to show the efficiency in fast JS distance computation of large real-world graphs, enabled by FINGER. Additionally, our ex post facto analysis shows high correlation of FINGER with an explainable anomaly proxy.
Beyond efficiency, we use the next two sets of experiments (with ground truths) to demonstrate the effectiveness of FINGER.

\textbf{Bifurcation detection results.} Using the ground-truth  statistic provided by \citep{liu2018genome},
we compare the performance of detecting the critical bifurcation point by each method. Let $\theta_{t,t^\prime}$ denote a dissimilarity metric between two graphs  $G_t$ and $G_{ t^\prime}$ from $\{G_t\}_{t=1}^T$.
For each method, the temporal difference score (TDS) proposed in \citep{liu2018genome} is used for bifurcation detection, which is defined as $\textnormal{TDS}(t)=\frac{1}{2}[\theta_{t, t-1}+\theta_{t, t+1}]$ when $t\in \{2,\ldots,T-1\}$, and  $\textnormal{TDS}(1)=\theta_{1,2}$ and $\textnormal{TDS}(T)=\theta_{T,T-1}$. The measurement(s) corresponding to a local minimum in TDS is detected as a bifurcation instance.
The ground-truth statistic and TDS of each method are shown in Figure \ref{Fig_app_cell}. Among all the compared methods, FINGER-JSdist (Algorithm \ref{algo_JSfast}) is the only method that correctly detects the bifurcation point (index 6), and its TDS based on JS distance also resembles the shape of the ground-truth statistic.

\textbf{Synthesized anomaly detection results.}
For further validation, we use another real-world dynamic peering network dataset at the autonomous system (AS) level (the Oregon-1 dataset \citep{leskovec2005graphs}) to synthesize anomalous connectivity patterns that mimic the denial-of-service (DoS) attacks. Here each graph represents the router connectivity over a certain time period, leading to 9 such graphs. We synthesize anomalous events by first selecting one graph from the first 8 graphs at random, and then connecting $X\%$ of nodes to a randomly chosen node in the selected graph. This synthesized connection pattern mimics that of the DoS attack, in which multiple nodes (e.g., a botnet) aim to connect to the target node simultaneously. The task is to detect this  synthesized  anomalous event by comparing the dissimilarity metric between consecutive graphs. 
Table \ref{table_oregon} reports the detection rate of different methods, where the detection rate is defined as the fraction of 100 random instances in which the anomalous event appears in the top-2 ranking based on a dissimilarity metric. Tested on $X=\{1,3,5,10\}\%$, FINGER-JS (Fast) consistently attains the best detection rate among all methods, suggesting the stability and superiority of the proposed method. 
On the other hand, the compared methods are not as robust as FINGER. Notably, when $X$ is small (i.e., the more challenging case for detection as the attack becomes stealthier), the detection performance of FINGER is more sensible than other methods. As $X$ becomes large, which means the DoS attack pattern is more apparent, the detection performance becomes similar.

\section{Conclusion}
In this paper, we proposed FINGER, a novel framework for efficiently computing von Neumann graph entropy (VNGE). FINGER reduces the computation of VNGE from cubic complexity to linear complexity for a given graph, and allows online computation based on incremental graph changes. In addition to bounded approximation error, our theory shows that FINGER is guaranteed to have asymptotic equivalence to the exact VNGE under mild conditions, which has been validated by extensive experiments on three different random graph models. The high  efficiency of FINGER also leads to scalable network learning algorithms for computing Jensen-Shannon distance between graphs.
Furthermore, we use two domain-specific applications and one synthesized dataset to corroborate the efficiency and effectiveness of FINGER compared to 7 baseline graph similarity methods. The results demonstrate the power  of FINGER in tackling large network analysis and (unsupervised) learning problems  in different domains. Our future work includes extension to directed graphs and negative edge weights.

\clearpage

\section*{Acknowledgment}
Pin-Yu Chen, Lingfei Wu and Sijia Liu acknowledge the support from MIT-IBM Watson AI Lab. Indika Rajapakse is supported in part by the Lifelong Learning Machines program from DARPA/MTO.

\bibliography{IEEEabrv,CPY_ref_20171004}
\bibliographystyle{icml2019}

\clearpage
\setcounter{equation}{0}
\setcounter{figure}{0}
\setcounter{table}{0}
\setcounter{page}{1}
\makeatletter
\renewcommand{\theequation}{S\arabic{equation}}
\renewcommand{\thefigure}{S\arabic{figure}}
\renewcommand{\thetable}{S\arabic{table}}
\section*{{ \LARGE Supplementary Material}}
\appendix

\section{Proof of Lemma \ref{lemma_quad_approx}}
\label{proof_lemma_quad_approx}
For any real $x$ such that $0<x<1$, it is easy to show that the Taylor series expansion of $-x\ln x$ at $1$ is  $\sum_{z=1}^\infty \frac{(-1)^z}{z} x(x-1)^z$. Applying this result to the term $- \lambda_i \ln \lambda_i $ in $H$ and taking the quadratic approximation of the series expansion gives
\begin{align}
\label{eqn_quad_approx_1}
Q=\sum_{i=1}^n \lambda_i(1 - \lambda_i)=1-\sum_{i=1}^n \lambda_i^2
\end{align}
since by definition $\sum_{i=1}^n\lambda_i=\trace(\bLNN)=1$.
The term $\sum_{i=1}^n \lambda_i^2$ in (\ref{eqn_quad_approx_1}) can be expressed as
\begin{align}
\label{eqn_quad_approx_2}
\sum_{i=1}^n \lambda_i^2&=\trace(\bLNN^2) \\
&=\sum_{i=1}^n \sum_{j=1}^n [\bLNN]_{ij} [\bLNN]_{ji}  \\
&\overset{(a)}{=}\sum_{i=1}^n \sum_{j=1}^n [\bLNN]_{ij}^2  \nonumber \\
&\overset{(b)}{=} c^2 \lb \sum_{i=1}^n  [\bL]_{ii}^2 + \sum_{i=1}^n \sum_{j=1, j\neq i}^n [\bL]_{ij}^2 \rb \\
&\overset{(c)}{=} c^2 \lb \sum_{i \in \cV} s_i^2 + 2  \sum_{(i,j) \in \cE} w_{ij}^2  \rb,
\end{align}
where $(a)$ is due to the matrix symmetry of $\bLNN$, $(b)$ is due to the definition that $\bLNN=c \cdot \bL$, and $(c)$ is due to the definition of $\bL$ such that $[\bL]_{ii}=s_i$, and $[\bL]_{ij}=w_{ij}$ when $(i,j) \in \cE$ and  $[\bL]_{ij}=0$ otherwise. Furthermore, define
\begin{align}
\label{eqn_quad_approx_3}
S=\trace(\bL)=\sum_{i=1}^n [\bL]_{ii}=\sum_{i \in \cV} s_i=2\sum_{(i,j)\in \cE} w_{ij}.
\end{align} 
Using the relation $c=\frac{1}{\trace(\bL)}$, we obtain the expression $Q=1-c^2 \lb \sum_{i \in \cV} s_i^2 + 2  \sum_{(i,j) \in \cE} w_{ij}^2  \rb$, where  $c=\frac{1}{S}$ and $S=\sum_{i \in V} s_i=2\sum_{(i,j)\in \cE} w_{ij}$.

\section{Proof of Theorem \ref{thm_entropy_bound}}
The assumption $\lambda_{\max} < 1 $ implies $0< \lambda_i \leq \lambda_{\max} < 1$ for all nonzero eigenvalues $\lambda_i$. Following the definition of $H$, we can rewrite $H$ as 
\begin{align}
\label{eqn_thm_entropy_bound_1}
H&=-\sum_{i=1}^n \lambda_i \ln \lambda_i \\
&=-\sum_{i: \lambda_i > 0} \lambda_i \ln \lambda_i \\
&=-\sum_{i: \lambda_i > 0} \lambda_i (1-\lambda_i) \frac{\ln \lambda_i}{1-\lambda_i}.
\label{eqn_thm_entropy_bound_1_c}
\end{align}
Since for all $\lambda_i > 0$, $\ln \lambda_{\min} \leq \ln \lambda_i \leq\ln \lambda_{\max}<0$ and $0<1- \lambda_{\max} \leq 1-\lambda_i \leq 1- \lambda_{\min} <1$, we obtain the relation
\begin{align}
\label{eqn_thm_entropy_bound_2}
\frac{-\ln \lambda_{\max}}{1-\lambda_{\min}} \leq \frac{-\ln \lambda_{i}}{1-\lambda_{i}} \leq  \frac{-\ln \lambda_{\min}}{1-\lambda_{\max}}.
\end{align}
Using $Q=\sum_{i=1}^n \lambda_i(1 - \lambda_i)=\sum_{i: \lambda_i>0} \lambda_i(1 - \lambda_i)$ in (\ref{eqn_quad_approx_1}) and applying (\ref{eqn_thm_entropy_bound_2})  to (\ref{eqn_thm_entropy_bound_1_c}) yields
\begin{align}
\label{eqn_thm_entropy_bound_3}
-Q \frac{\ln \lambda_{\max}}{1-\lambda_{\min}} \leq H \leq -Q \frac{\ln \lambda_{\min}}{1-\lambda_{\max}}.
\end{align}
When $G$ is a complete graph with identical edge weight $x>0$, it can be shown that the eigenvalues of $\bL$ have 1  eigenvalue at $0$ and $n-1$ identical eigenvalues at $nx$ \citep{Merris94}. Since the trace normalization constant $c=\frac{1}{\trace(\bL)}=\frac{1}{(n-1)nx}$, the eigenvalues of $\bLNN=c \cdot \bL$ are $\lambda_n=0$ and $\lambda_{i}=\frac{nx}{(n-1)nx}=\frac{1}{n-1}$ for all $1 \leq i \leq n-1$, which implies $H=\ln(n-1)$. It is easy to see that  in this case $Q=1- \frac{1}{n-1}=1-\lambda_{\min}=1-\lambda_{\max}$ and $-\ln \lambda_{\max}=-\ln \lambda_{\min}=\ln(n-1)$. Consequently, the bounds in (\ref{eqn_thm_entropy_bound_3}) become exact and $H=\ln (n-1)$ when $G$ is a complete graph with identical edge weight.

\section{On the condition $\lambda_{\max}<1$ in Theorem \ref{thm_entropy_bound}}
Here we show that the condition $\lambda_{\max}<1$ is always satisfied with any graph $G \in \cG$ having a connected subgraph with at least 3 nodes. By definition, $\lambda_{\max} \leq 1$ since it is the largest eigenvalue of the scaled matrix $\bLNN= \bL/\trace(\bL)$. 
Since any connected subgraph  with at least 3 nodes will contribute to at least 2 positive eigenvalues of $\bLNN$ \citep{Mieghem10,CPY13GlobalSIP} and all eigenvalues of $\bLNN$ sum to 1, we have $\lambda_{\max} < 1$.

\section{Proof of Corollary \ref{cor_asymptotic_Q}}
Since $\sum_{i=1}^n \lambda_i=1$, the condition $\lambda_{\min} = \Omega(\lambda_{\max}) $ implies $\lambda_{\max}$ and $\lambda_{\min}$ are of the same order $\frac{1}{n_+}$, where $n_+$ is the number of positive eigenvalues of $\bLNN$. When the condition $n_+ = \Omega (n)$ also holds, then  $\lambda_{\max}=\frac{a}{n}$ and  $\lambda_{\min}=\frac{b}{n}$ for some constants $a,b$ such that $a \geq b >0$, and we obtain
\begin{align}
\label{eqn_cor_asymptotic_Q_1}
\lim_{n \ra \infty}-\frac{1}{\ln n} \cdot \frac{\ln \lambda_{\max}}{1-\lambda_{\min}}= \lim_{n \ra \infty} \frac{1}{\ln n}  \cdot \frac{\ln n - \ln a}{1-\frac{b}{n}} = 1.
\end{align}
Similarly, 
\begin{align}
\label{eqn_cor_asymptotic_Q_2}
\lim_{n \ra \infty}-\frac{1}{\ln n} \cdot \frac{\ln \lambda_{\min}}{1-\lambda_{\max}} = 1.
\end{align}
Taking the limit of $\frac{H}{ \ln n}$ and applying (\ref{eqn_cor_asymptotic_Q_1}) and (\ref{eqn_cor_asymptotic_Q_2}) to the bounds in (\ref{eqn_thm_entropy_bound_3}), we obtain
\begin{align}
\label{eqn_cor_asymptotic_Q_3}
\lim_{n \ra \infty} \frac{H}{\ln n} - Q= 0,
\end{align}
which completes the proof.

\section{Proof of Corollary \ref{cor_approx_error_hH}}
Following the proof of Corollary \ref{cor_asymptotic_Q}, if  $n_+ = \Omega(n)$ and $\lambda_{\min} = \Omega(\lambda_{\max})$, then $\lambda_{\max}=\frac{a}{n}$ and  $\lambda_{\min}=\frac{b}{n}$ for some constants $a,b$ such that $a \geq b >0$. We have
\begin{align}
\label{eqn_cor_approx_error_hH_1}
\lim_{n \ra \infty}\frac{H-\hH}{\ln n}&= \lim_{n \ra \infty} \frac{H}{\ln n} - Q + Q - \frac{\hH}{\ln n} \\
&\overset{(a)}{=}\lim_{n \ra \infty} Q - \frac{\hH}{\ln n}  \\
&\overset{(b)}{=}\lim_{n \ra \infty} Q - Q \cdot \frac{\ln n - \ln a}{\ln n} \\
&=0,
\end{align}
where $(a)$ uses (\ref{eqn_cor_asymptotic_Q_3}) and $(b)$ uses the definition of $\hH$ in (\ref{eqn_H_hat}) and $\lambda_{\max}=\frac{a}{n}$. This implies the approximation error $H-\hH$ decays with $\ln n$. That is, $H-\hH=o(\ln n)$.

\section{Proof of Corollary \ref{cor_approx_error_tH}}
Let $\mu_{\max}$ denote the largest eigenvalue of the graph Laplacian matrix $\bL$ of a graph $G \in \cG$. Then it is known that 
$\frac{n}{n-1} s_{\max} \leq \mu_{\max} \leq 2 s_{\max}$, where the lower bound is proved in \citep{Fiedler73} and the upper bound is proved in \citep{anderson1985eigenvalues}. These bounds suggest that $\mu_{\max}$ has asymptotically the same order as $s_{\max}$. Moreover, since by definition $\bLNN = c \cdot \bL$, it holds that $\lambda_{\max}= c \cdot \mu_{\max}$ and hence $\lambda_{\max} = O ( c \cdot s_{\max} )$. Following the proof of Corollary \ref{cor_asymptotic_Q}, if  $n_+ =\Omega( n)$ and $\lambda_{\min} = \Omega(\lambda_{\max})$, then $\lambda_{\max}=\frac{a}{n}$ and  $\lambda_{\min}=\frac{b}{n}$ for some constants $a,b$ such that $a \geq b >0$, and $2 c \cdot s_{\max}=\frac{\gamma}{n}$ for some $\gamma>0$ since $\lambda_{\max}= O( c \cdot s_{\max} )$. Similar to the proof of Corollary \ref{cor_approx_error_hH}, we have
\begin{align}
\label{eqn_cor_approx_error_tH_1}
\lim_{n \ra \infty}\frac{H-\tH}{\ln n}&= \lim_{n \ra \infty} \frac{H}{\ln n} - Q + Q - \frac{\tH}{\ln n} \\
&\overset{(a)}{=}\lim_{n \ra \infty} Q - \frac{\tH}{\ln n}  \\
&\overset{(b)}{=}\lim_{n \ra \infty} Q - Q \cdot \frac{\ln n - \ln \gamma}{\ln n} \\
&=0,
\end{align}
where $(a)$ uses (\ref{eqn_cor_asymptotic_Q_3}) and $(b)$ uses the definition of $\tH$ in (\ref{eqn_tH}) and  $2 c \cdot s_{\max}=\frac{\gamma}{n}$. This implies the approximation error $H-\tH$ decays with $\ln n$. That is, $H-\tH=o(\ln n)$.

\begin{figure*}[t]
	\centering
	\begin{subfigure}[b]{0.5\linewidth}
		\includegraphics[width=\textwidth]{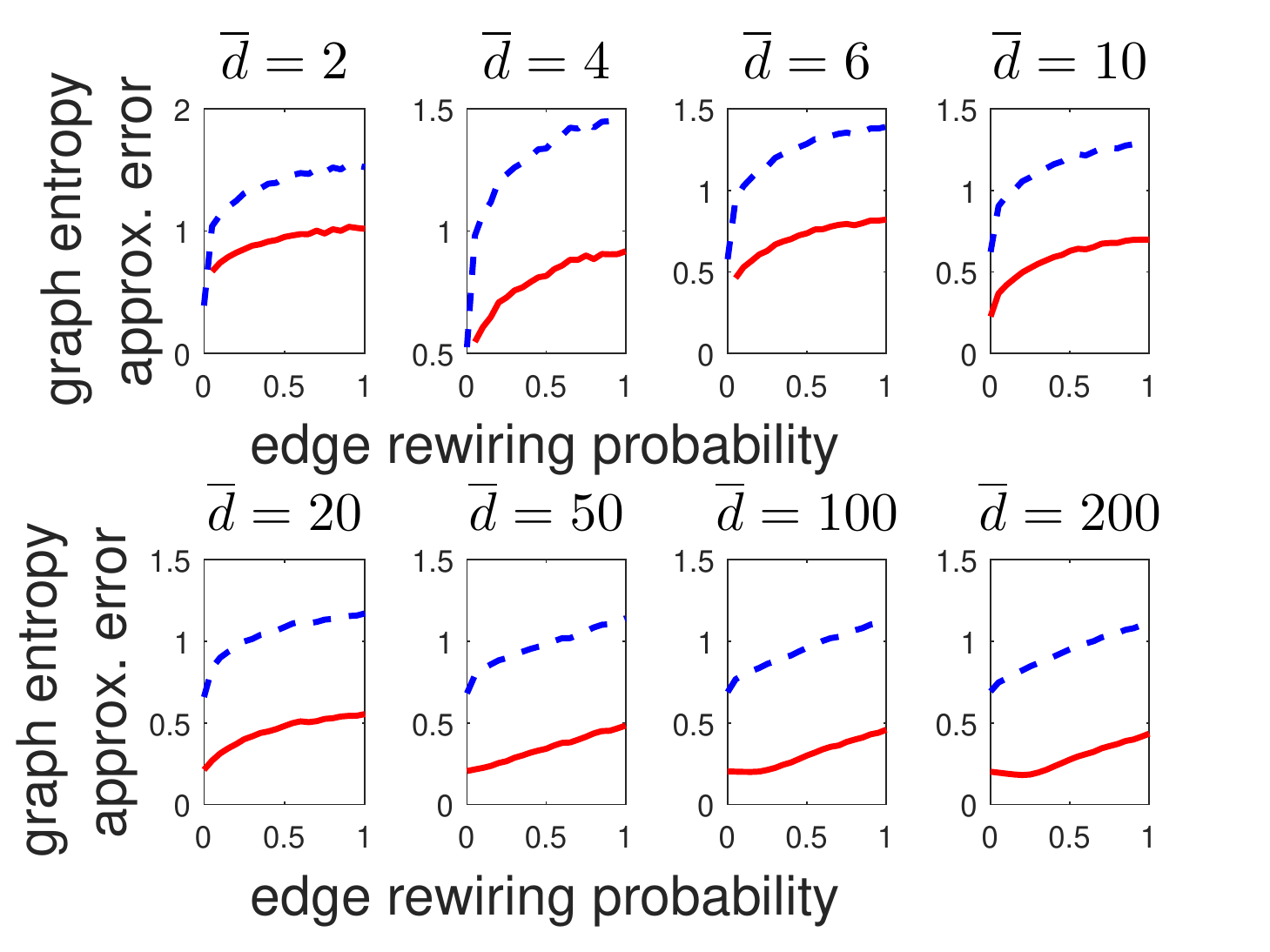}
		\caption{Approximation error}
	\end{subfigure}%
	\centering
	\begin{subfigure}[b]{0.5\linewidth}
		\includegraphics[width=\textwidth]{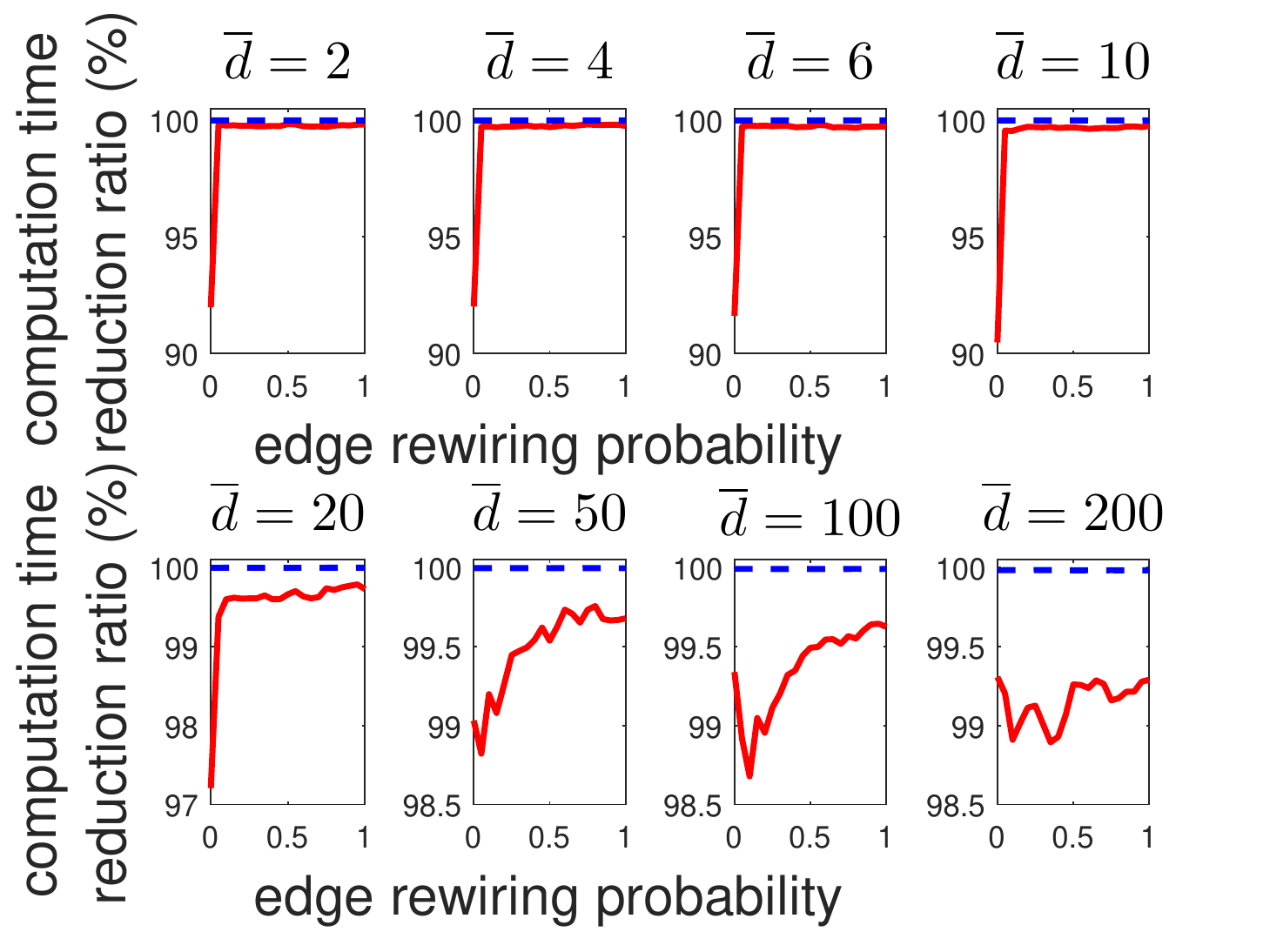}
		\caption{CTRR}
	\end{subfigure}
	\caption{Approximation error and computation time reduction ratio (CTRR) of FINGER  under different average degree $\od$ of WS model. The red solid  line and  blue dashed line refer to the results of $\hH$ and $\tH$, respectively. Both $\hH$ and $\tH$ achieve at least 97\% speed-up relative to the computation of $H$ in all cases.  It is observed that $\tH$ has larger approximation error than $\hH$ but better CTRR. }
	\label{Fig_FINGER_WS_all}
\end{figure*}

\section{Proof of Theorem \ref{thm_incremental}}
Let $\bL$ and $\bpL$ denote the graph Laplacian matrix of $G$ and $\pG$, respectively, and let $\bLNN= c \cdot \bL$ and  $\bpLN=\pc \cdot \bpL$ be the corresponding trace-normalized matrices. Since $S=\trace(\bL)=2 \sum_{(i,j) \in \cE} w_{ij}$ and $\Delta S=2 \sum_{(i,j) \in \Delta \cE} \Delta w_{ij}$, it is easy to show that $\trace(\bpL)=S+\Delta S=1/\pc$.  We have
\begin{align}
\label{eqn_thm_incremental_1}
\pc - c= \frac{1}{S + \Delta S} - \frac{1}{S}=\frac{-\Delta S}{(S + \Delta S) S} = -c \pc \Delta S
\end{align}
since $\pc=1/\trace{(\bpL)}$ and $c=1/\trace{(\bL)}$.
This then implies $\pc=\frac{c}{1+c \Delta S}$ and 
\begin{align}
\label{eqn_thm_incremental_2}
\Delta c = \pc - c = \frac{-c^2 \Delta S}{1+c \Delta S}.
\end{align}
Using the expression of quadratic approximation for VNGE in Lemma \ref{lemma_quad_approx} and the relation that $\pG=G \oplus \Delta G$, we have
\begin{align}
\label{eqn_thm_incremental_3}
&Q - \pQ \nonumber  \\
&= (c+ \Delta c)^2 \lb \sum_{i \in \cV}  (s_i + \Delta s_i)^2 + 2 \sum_{(i,j) \in \cE}  (w_{ij}+ \Delta w_{ij})^2 \rb \nonumber \\
&~~~- c^2 \lb \sum_{i \in \cV}  s_i ^2 + 2 \sum_{(i,j) \in \cE}  w_{ij}^2 \rb \\
\label{eqn_thm_incremental_3_1}
&=( 2 \Delta c + \Delta c^2) \lb \sum_{i \in \cV} s_i^2 + 2 \sum_{(i,j)\in \cE} w_{ij}^2 + \Delta Q \rb \nonumber  \\ 
&~~~+ c^2 \Delta Q,
\end{align}
where $\Delta Q = 2 \sum_{i \in \Delta \cV } s_i \Delta s_i + \sum_{i \in \Delta \cV }  \Delta s_i^2 + 4 \sum_{(i,j) \in \Delta \cE} w_{ij} \Delta w_{ij} + 2\sum_{(i,j) \in \Delta \cE} \Delta w_{ij}^2 $, and we use the convention $\Delta s_i =0$ and $\Delta w_{ij}=0$ when there are no changes made in the nodal strength of node $i$ and in the weight of edge $(i,j)$ from $G$ to $\pG$, respectively. Since $Q= 1- c^2 \lb \sum_{i \in \cV} s_i^2 + 2 \sum_{(i,j)\in \cE} w_{ij}^2  \rb$, replacing $\sum_{i \in \cV} s_i^2 + 2 \sum_{(i,j)\in \cE} w_{ij}^2 $ with $\frac{1-Q}{c^2}$ in (\ref{eqn_thm_incremental_3_1}) and using the relation $\pc=c+\Delta c$  yields
\begin{align}
\label{eqn_thm_incremental_4}
\pQ = \lb \frac{\pc}{c} \rb ^2 Q - \pc^2 \Delta Q - \frac{2 \Delta c + \Delta c^2}{c^2}. 
\end{align}
Using the result from (\ref{eqn_thm_incremental_2}) that $ \frac{\pc}{c} = \frac{1}{1+c \Delta S} $, we can further simplify (\ref{eqn_thm_incremental_4}) as
\begin{align}
\label{eqn_thm_incremental_5}
\pQ &= \frac{Q}{(1+ c \Delta S )^2}    - \lb \frac{c}{1+c \Delta S} \rb^2 \Delta Q -  \frac{1}{\lb 1+c \Delta S\rb^2}  +1   \\
&= \frac{Q-1}{(1+ c \Delta S )^2}   - \lb \frac{c}{1+c \Delta S} \rb^2 \Delta Q   +1,
\end{align}
which completes the proof.

\begin{figure*}[t]
	\centering
	\begin{subfigure}[b]{0.42\linewidth}
		\includegraphics[width=\textwidth]{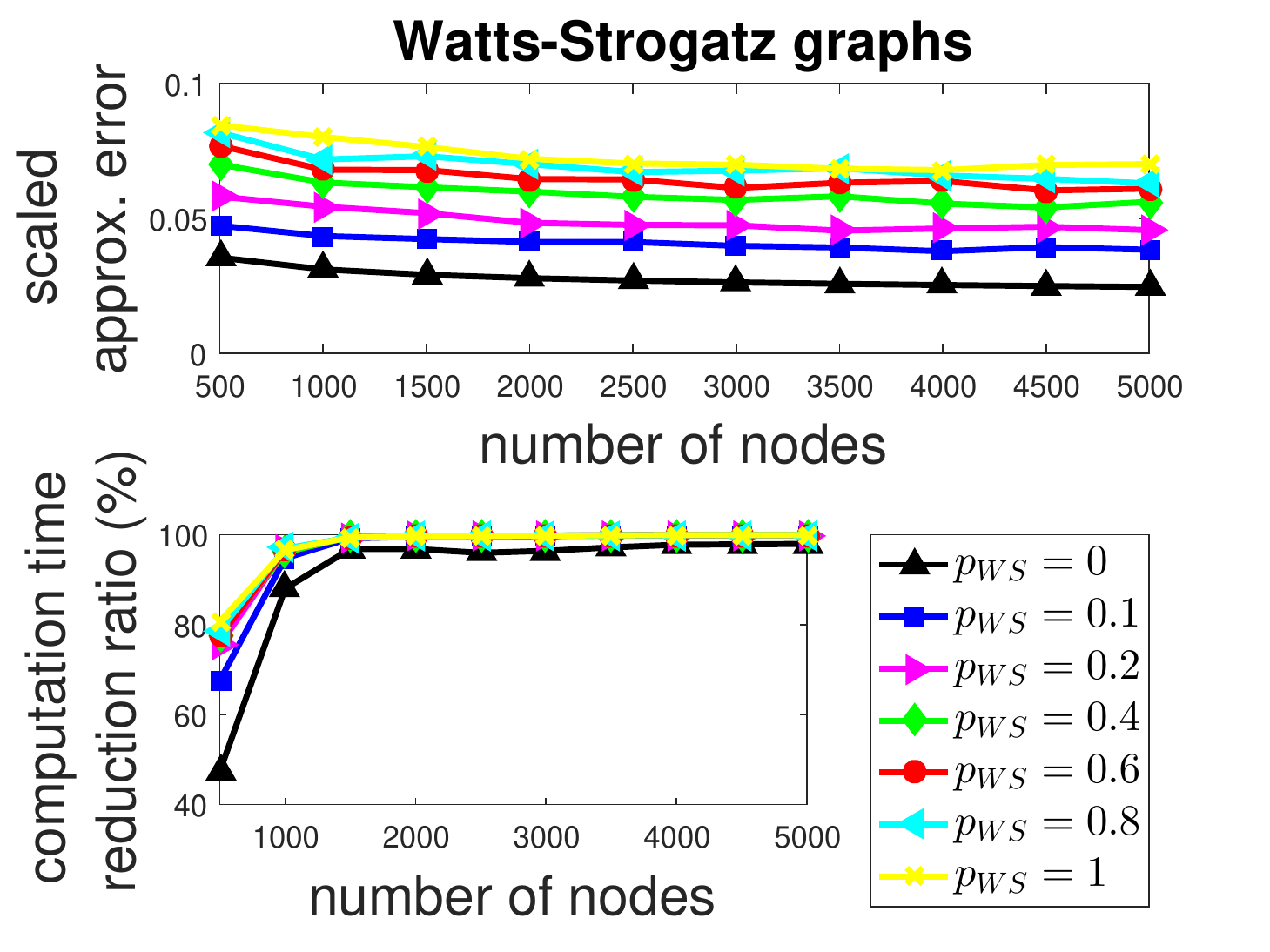}
		\caption{WS model ($\od=20$)}
	\end{subfigure}		
	\begin{subfigure}[b]{0.42\linewidth}
		\centering
		\includegraphics[width=\textwidth]{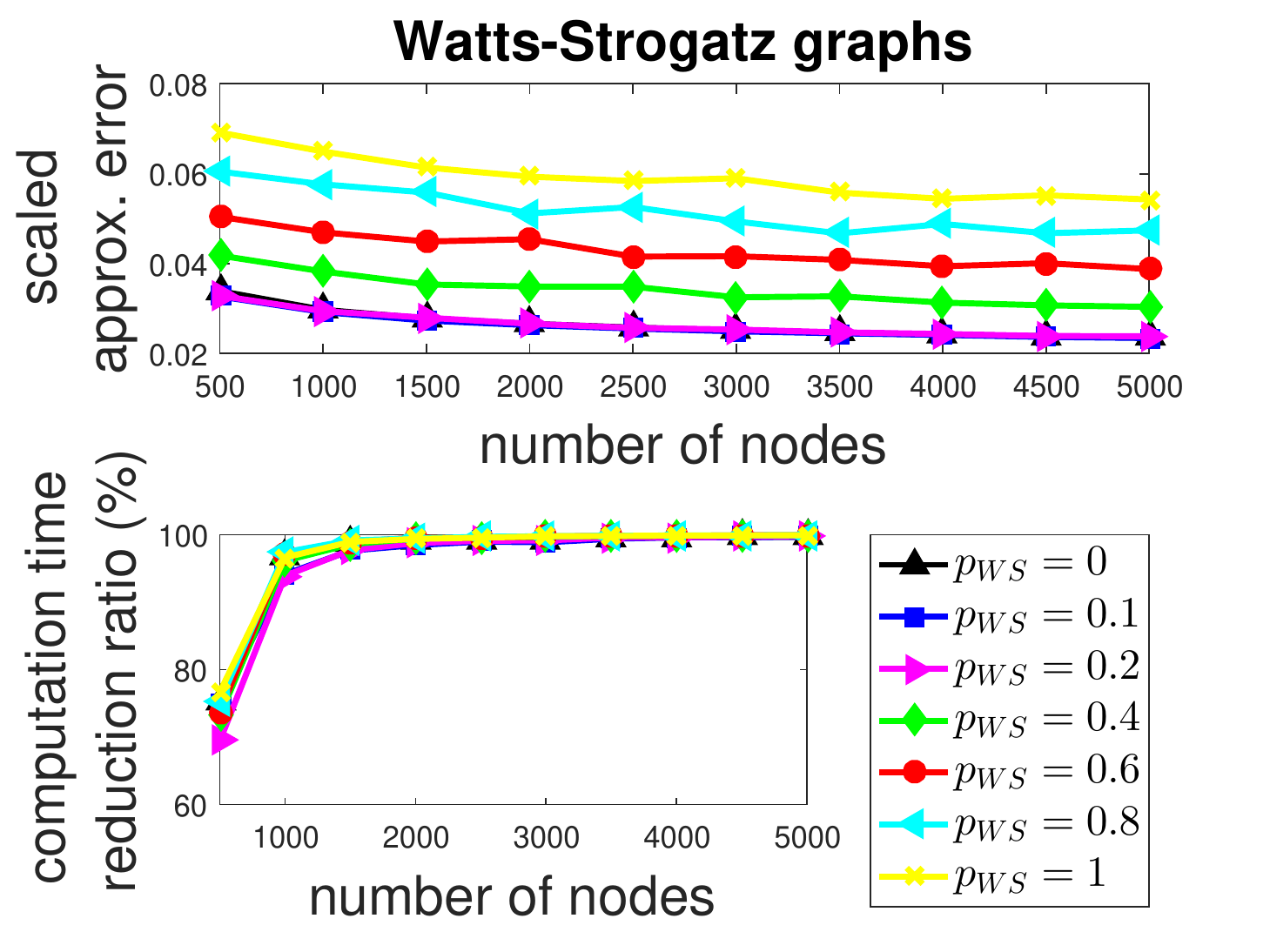}
		\caption{WS model ($\od=100$)}
	\end{subfigure}			
	\caption{Scaled approximation error (SAE) and computation time reduction ratio (CTRR) of $\hH$ via FINGER for WS model under varying number of nodes $n$. Their behaviors are similar to the case of $\od=50$ as displayed in Figure \ref{Fig_FINGER_node} (c).}
	\label{Fig_FINGER_node_WS_hH}
\end{figure*}

\begin{figure*}[t!]
	\centering
	\begin{subfigure}[b]{0.42\linewidth}
		\includegraphics[width=\textwidth]{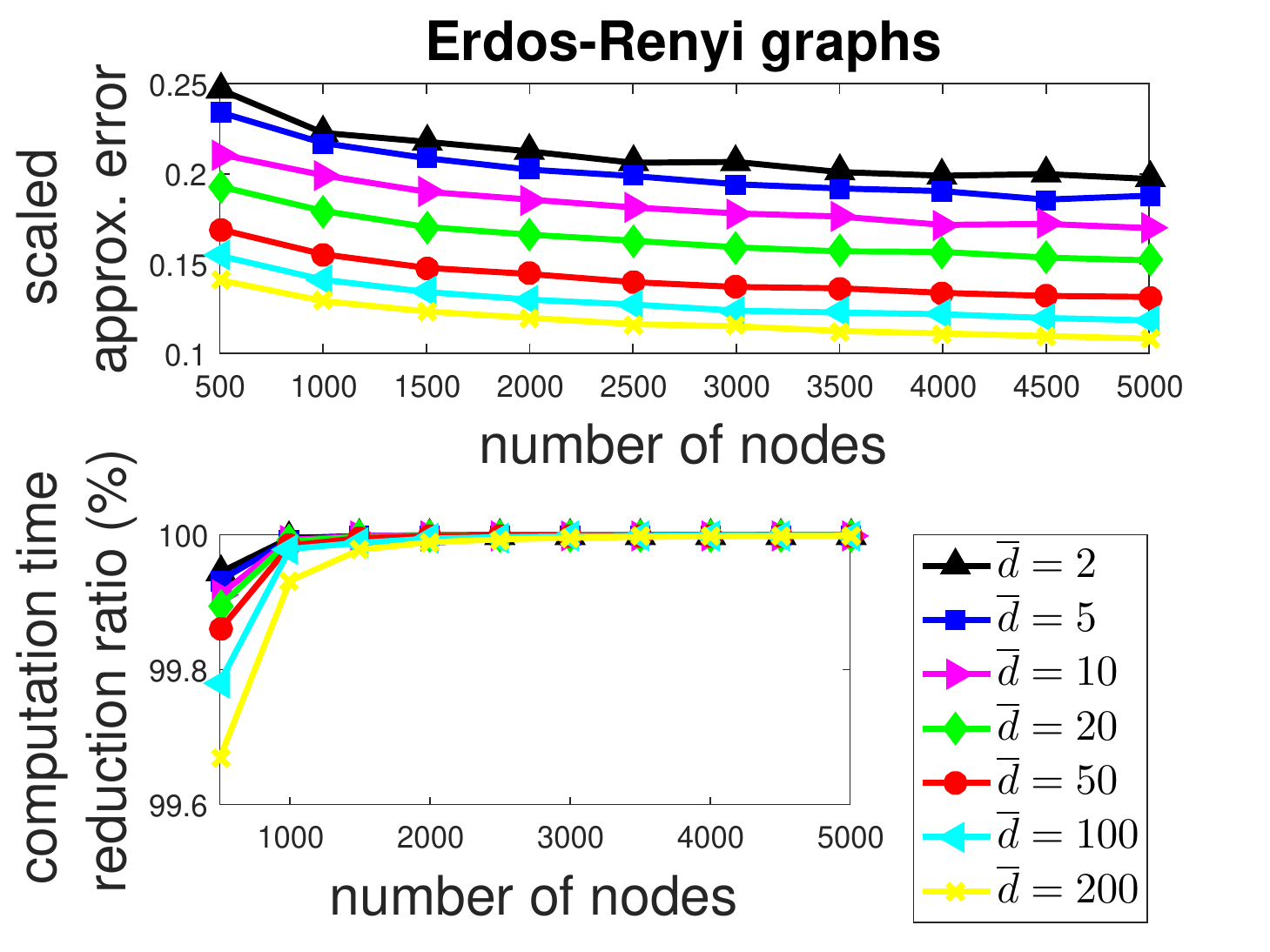}
		\caption{ER model}
	\end{subfigure}%
	\centering
	\begin{subfigure}[b]{0.42\linewidth}
		\includegraphics[width=\textwidth]{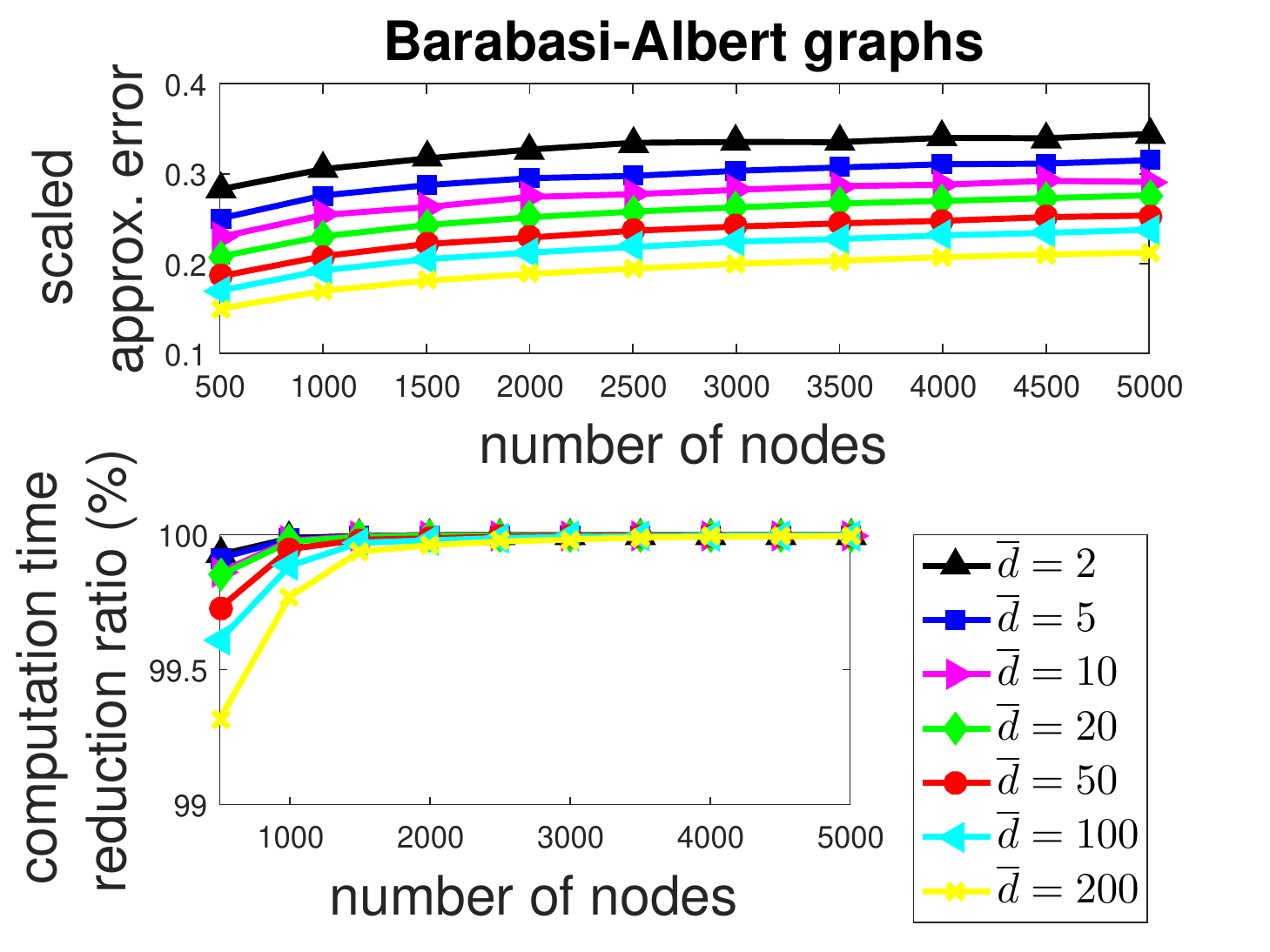}
		\caption{BA model}
	\end{subfigure}
	\\
	\begin{subfigure}[b]{0.42\linewidth}
		\includegraphics[width=\textwidth]{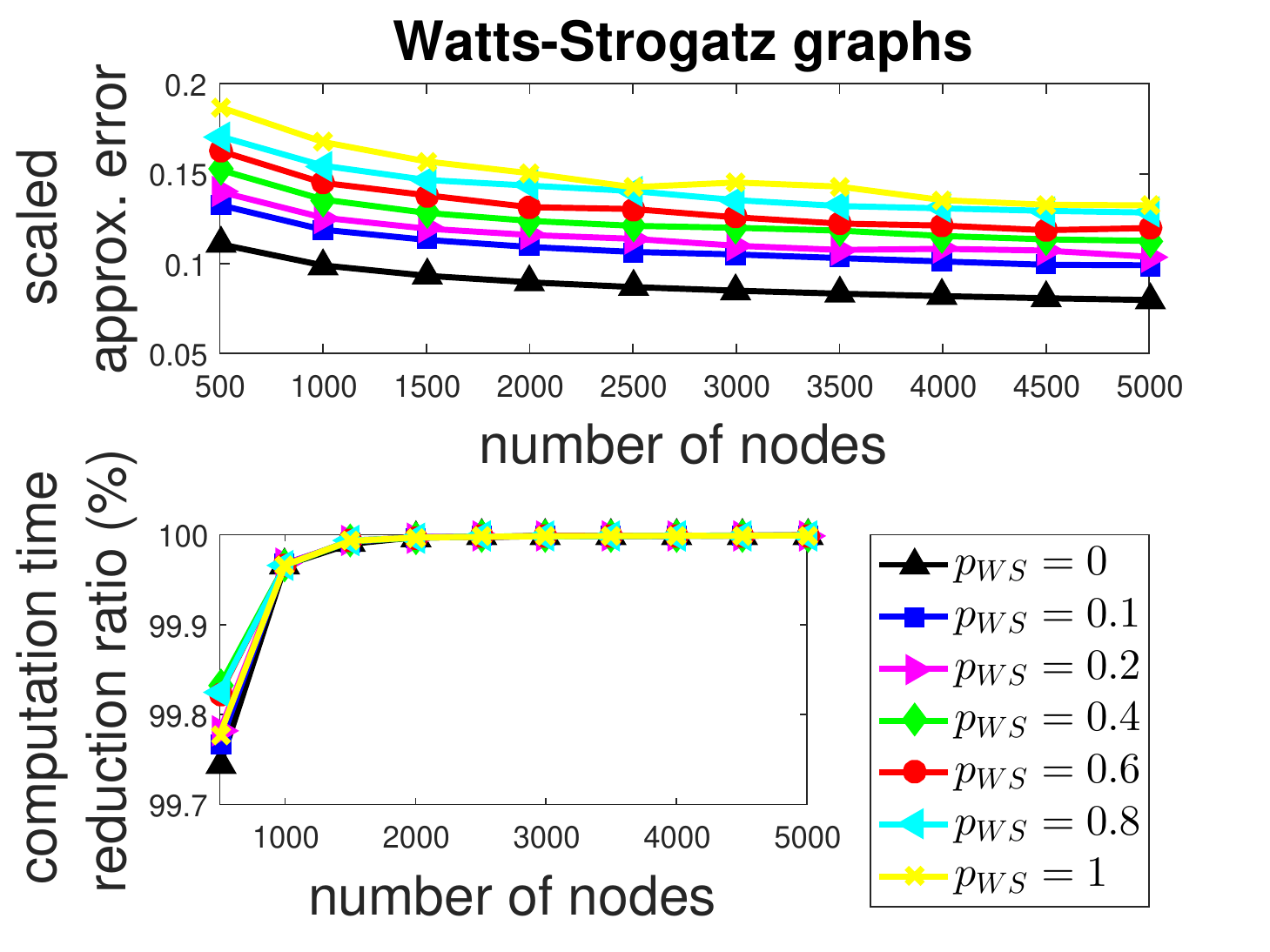}
		\caption{WS model ($\od=50$)}
	\end{subfigure}		
	\begin{subfigure}[b]{0.42\linewidth}
		\includegraphics[width=\textwidth]{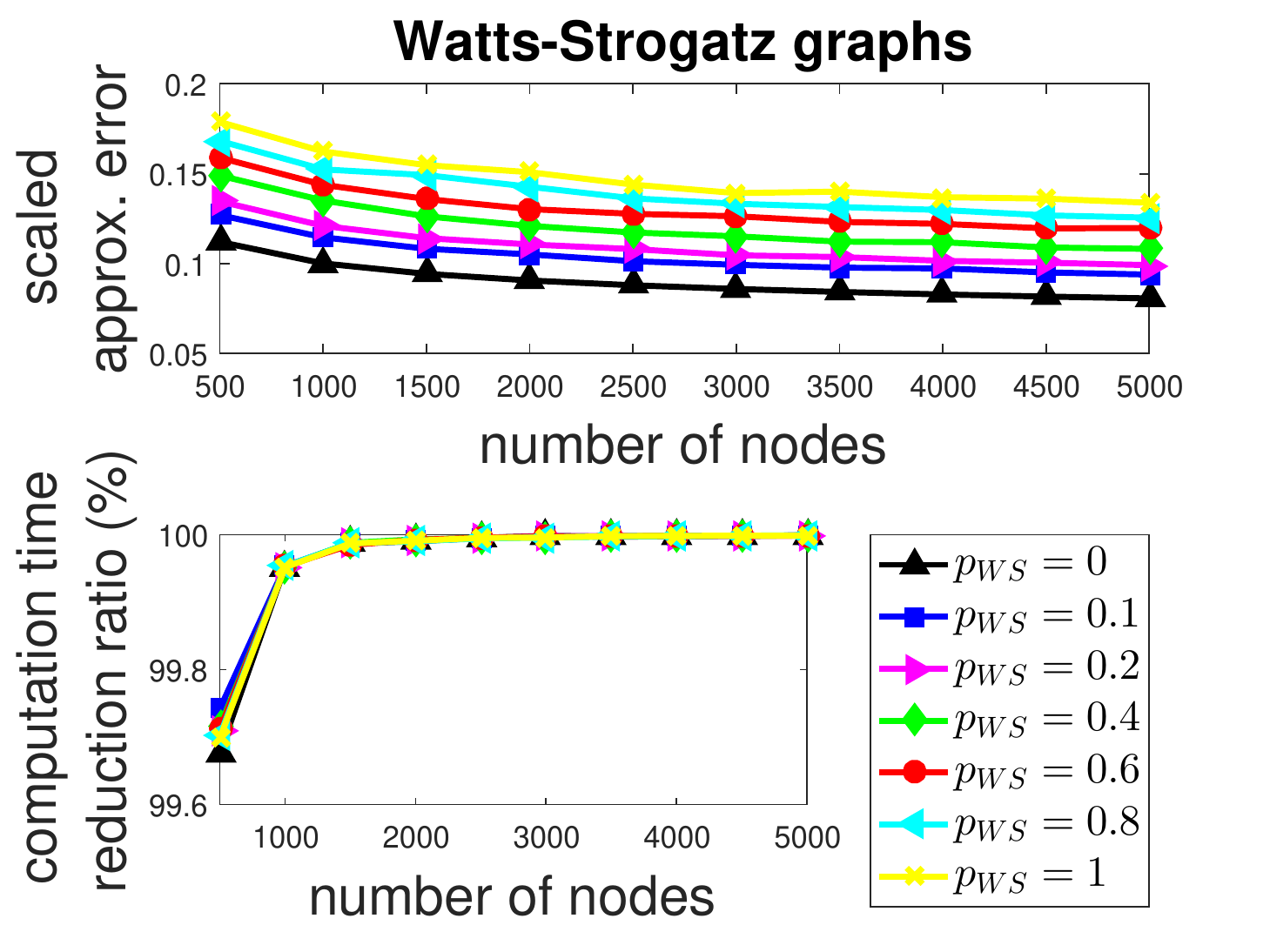}
		\caption{WS model ($\od=100$)}
	\end{subfigure}			
	\caption{Scaled approximation error (SAE) and computation time reduction ratio (CTRR) of $\tH$ via FINGER for different random graph models and varying number of nodes $n$. The SAE of ER and WS graphs validates the $o(\ln n)$ approximation error analysis in  Corollary \ref{cor_approx_error_tH}, whereas the SAE of BA graphs grows logarithmically with $n$ due to imbalanced eigenspectrum. The CTRR attains nearly 100\% speed-up relative to $H$ for  moderate-size graphs ($n \geq 1500$).}
	\label{Fig_FINGER_node_tH}
\end{figure*}

\section{Finite-size analysis and asymptotic equivalence of JS distance using FINGER}
Beyond asymptotic analysis, we believe our results can provide new insights to finite-size analysis, especially based on the facts that: (i) our entropy inequality $\tH \leq \hH \leq H$ is a finite-size result; (ii) The VNGE approximation error rate $o(\ln n)$ is in fact optimal in $n$ for any finite-size analysis, since Theorem \ref{thm_entropy_bound} shows that the rate is tight for complete graphs with identical edge weights.

Furthermore, based on the asymptotic equivalence results of VNGE, it is straightforward to establish asymptotic equivalence of JS distance using FINGER as described in Algorithms \ref{algo_JSfast} and \ref{algo_JSinc}.
Let $\textsf{JS}$ denote the exact JS distance and $\textsf{JS}_{\textnormal{FINGER}}$ denote the approximate JS distance using the VNGE computation from FINGER (either $\hH$ or $\tH$). 
Using Corollaries 2 and 3, the properly scaled absolute approximation error (SAAE) of JS distance, $\frac{|\textsf{JS}-\textsf{JS}_{\textnormal{FINGER}}|}{\sqrt{\ln n}}$, converges to $0$ as $n \ra \infty$, which proves $|\textsf{JS}-\textsf{JS}_{\textnormal{FINGER}}|=o(\sqrt{\ln n})$ and $\frac{\textsf{JS}_{\textnormal{FINGER}}} { \sqrt{\ln n}}$ is asymptotically a distance metric.


\section{Additional experimental results on synthetic random graphs}
\paragraph{The effect of average degree $\od$ on Watts-Strogatz graphs.} 
Figure \ref{Fig_FINGER_WS_all} displays the approximation error and 
computation time reduction ratio (CTRR) of FINGER-$\hH$ and FINGER-$\tH$ under different average degree $\od$ of WS model, which is defined as $H-\hH$ and $H-\tH$, respectively. It can be observed that when fixing $\od$, the approximation error decays with the edge rewiring probability for both $\hH$ and $\tH$. In addition, for the same  edge rewiring probability, larger $\od$ yields less approximation error. Using FINGER, both $\hH$ and $\tH$ achieve at least 97\% speed-up relative to the computation of $H$ in all cases. The approximate VNGE $\tH$ always attains  better CTRR than $\hH$ but at the price of larger approximation error due to the fact that $\tH \leq \hH \leq H$.

Figure \ref{Fig_FINGER_node_WS_hH} displays the scaled approximation error (SAE) and computation time reduction ratio of $\hH$ via FINGER for WS model under varying number of nodes $n$ and two different settings of the average degree $\od$.  Their behaviors are similar to the case of $\od=50$ as displayed in  Figure \ref{Fig_FINGER_node} (c). 

\begin{figure*}[t]
	\centering
	\begin{subfigure}[b]{0.48\textwidth}
		\includegraphics[width=\textwidth]{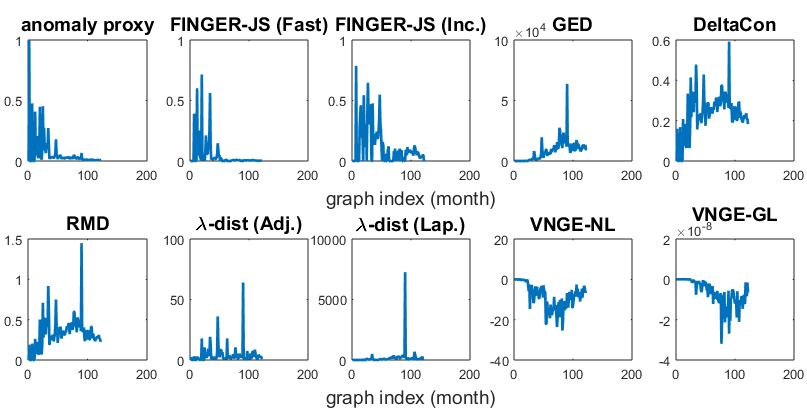}
		\caption{Dissimilarity (anomaly) metrics of Wikipedia-sEN}
	\end{subfigure}%
	\hspace{5mm}
	\centering
	\begin{subfigure}[b]{0.47\textwidth}
		\includegraphics[width=\textwidth]{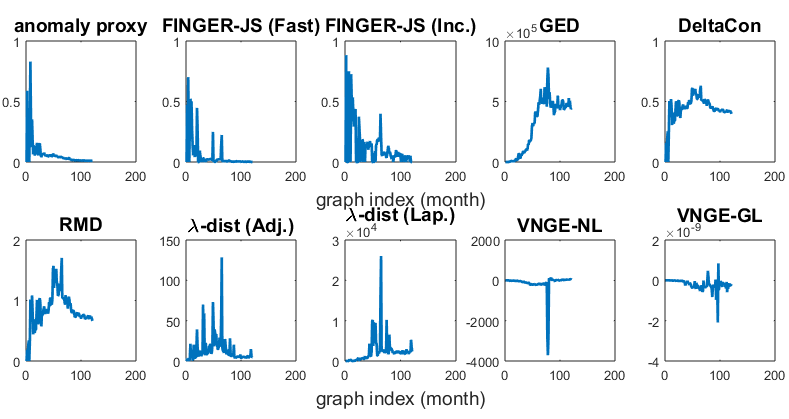}
		\caption{Dissimilarity (anomaly) metrics of Wikipedia-FR}
	\end{subfigure}
	\\
	\vspace{3mm}
	\centering
	\begin{subfigure}[b]{0.47\textwidth}
		\includegraphics[width=\textwidth]{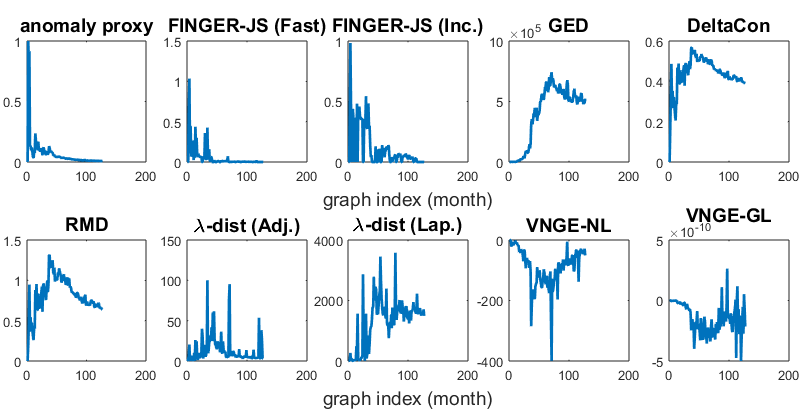}
		\caption{Dissimilarity (anomaly) metrics of Wikipedia-GE}
	\end{subfigure}		
	\caption{Anomaly detection in consecutive monthly Wikipedia hyperlink networks via different 
		dissimilarity metrics. The corresponding computation time and Pearson correlation coefficient are reported in Table \ref{table_wiki_PCC}. Similar to the observations in Figure \ref{Fig_app_wiki}, FINGER-JSdist (Fast) best aligns with the anomaly proxy in all datasets.  FINGER-JSdist (Incremental) has efficient computation time but less consistency (second best PCC among all methods).}	
	\label{Fig_app_Wiki_other}	
	\vspace{-3mm}
\end{figure*}

\paragraph{The effect of graph size $n$ on FINGER-$\tH$.} 

In comparison to $\hH$ via FINGER in Figure \ref{Fig_FINGER_node}, Figure \ref{Fig_FINGER_node_tH} displays the SAE and CTRR of $\tH$ for the three different random graph models and varying number of nodes $n$. Consistent with the findings in Section \ref{sec_experiemnts},
the SAE of $\tH$ for ER and WS graphs obeys the $o(\ln n)$ approximation error analysis as established in  Corollary \ref{cor_approx_error_tH} since they have balanced eigenspectrum. On the other hand, the SAE of BA graphs grows logarithmically with $n$ due to imbalanced eigenspectrum. Fixing $n$, larger average degree or more graph regularity leads to less approximation error.
Comparing to $\hH$, the CTRR of $\tH$ attains nearly 100\% speed-up relative to $H$ for  relatively small-size graphs ($n \geq 1500$).

\begin{table*}[t!]
	\centering
	\caption{Performance comparison of Spearman's rank correlation coefficient (SRCC) between the anomaly proxy and each method in the Wikipedia application. FINGER attains the best SRCC across all datasets.}
	\label{table_SRCC}
	\begin{adjustbox}{max width=0.9\textwidth}		
		\begin{tabular}{llllllllll}
			\hline
			\multicolumn{1}{c}{Datasets}                         & \begin{tabular}[c]{@{}l@{}}FINGER\\ -JS (Fast)\end{tabular} & \begin{tabular}[c]{@{}l@{}}FINGER\\ -JS (Inc.)\end{tabular} & DeltaCon & RMD     & \begin{tabular}[c]{@{}l@{}}$\lambda$ dist. \\ (Adj.)\end{tabular} & \begin{tabular}[c]{@{}l@{}}$\lambda$ dist. \\ (Lap.)\end{tabular} & GED     & \begin{tabular}[c]{@{}l@{}}VNGE\\ -NL\end{tabular} & \begin{tabular}[c]{@{}l@{}}VNGE\\ -GL\end{tabular} \\ \hline
			\begin{tabular}[c]{@{}l@{}}Wiki\\ (sEN)\end{tabular} & \textbf{0.5055}                                             & 0.3849                                                      & 0.4518   & 0.4518  & 0.4208                                                            & 0.0402                                                            & -0.1355 & -0.0542                                            & 0.2231                                             \\ \hline
			\begin{tabular}[c]{@{}l@{}}Wiki\\ (EN)\end{tabular}  & \textbf{0.7973}                                             & 0.5039                                                      & -0.4620  & -0.4620 & -0.3014                                                           & -0.5981                                                           & -0.7759 & -0.1823                                            & 0.4840                                             \\ \hline
			\begin{tabular}[c]{@{}l@{}}Wiki\\ (FR)\end{tabular}  & \textbf{0.7026}                                             & 0.4563                                                      & 0.2652   & 0.2652  & 0.4297                                                            & -0.4355                                                           & -0.6125 & -0.4792                                            & 0.3938                                             \\ \hline
			\begin{tabular}[c]{@{}l@{}}Wiki\\ (GE)\end{tabular}  & \textbf{0.6591}                                             & 0.4930                                                      & 0.3167   & 0.3167  & 0.3707                                                            & -0.4343                                                           & -0.5695 & -0.0156                                            & 0.2606                                             \\ \hline
		\end{tabular}
	\end{adjustbox}
\end{table*}

\section{Implementation details for VNGE-NL and VNGE-GL}
We note that in the Wikipedia application, we omit the edge direction for all  methods except VNGE-GL since the resulting performance is almost identical. The implementation of VNGE-GL indeed considers the edge direction. We also note that in these two applications, the Jensen-Shannon distances of VNGE-NL and VNGE-GL are ineffective. Therefore, we use the consecutive difference of their approximate VNGE as the anomaly score, and take the absolute value of the anomaly score for anomaly ranking.


\section{Additional results for anomaly detection in evolving Wikipedia hyperlink networks}
\textbf{Additional Wikipedia network plots.}
The plots of dissimilarity (anomaly) metrics of different methods in Section \ref{sec_app} for consecutive monthly hyperlink networks  of
Wikipedia-sEN, Wikipedia-FR, and Wikipedia-GE are  
shown in Figure \ref{Fig_app_Wiki_other}. Their performance in terms of the computation time and Pearson correlation coefficient are reported in Table \ref{table_wiki_PCC}. Similar to the observations in Figure \ref{Fig_app_wiki}, FINGER-JSdist (Fast) best aligns with the anomaly proxy in all datasets.  FINGER-JSdist (Incremental) has efficient computation time but less consistency (still attains second best PCC among all methods).

\begin{figure*}[t]
	\centering
	\includegraphics[width=6in]{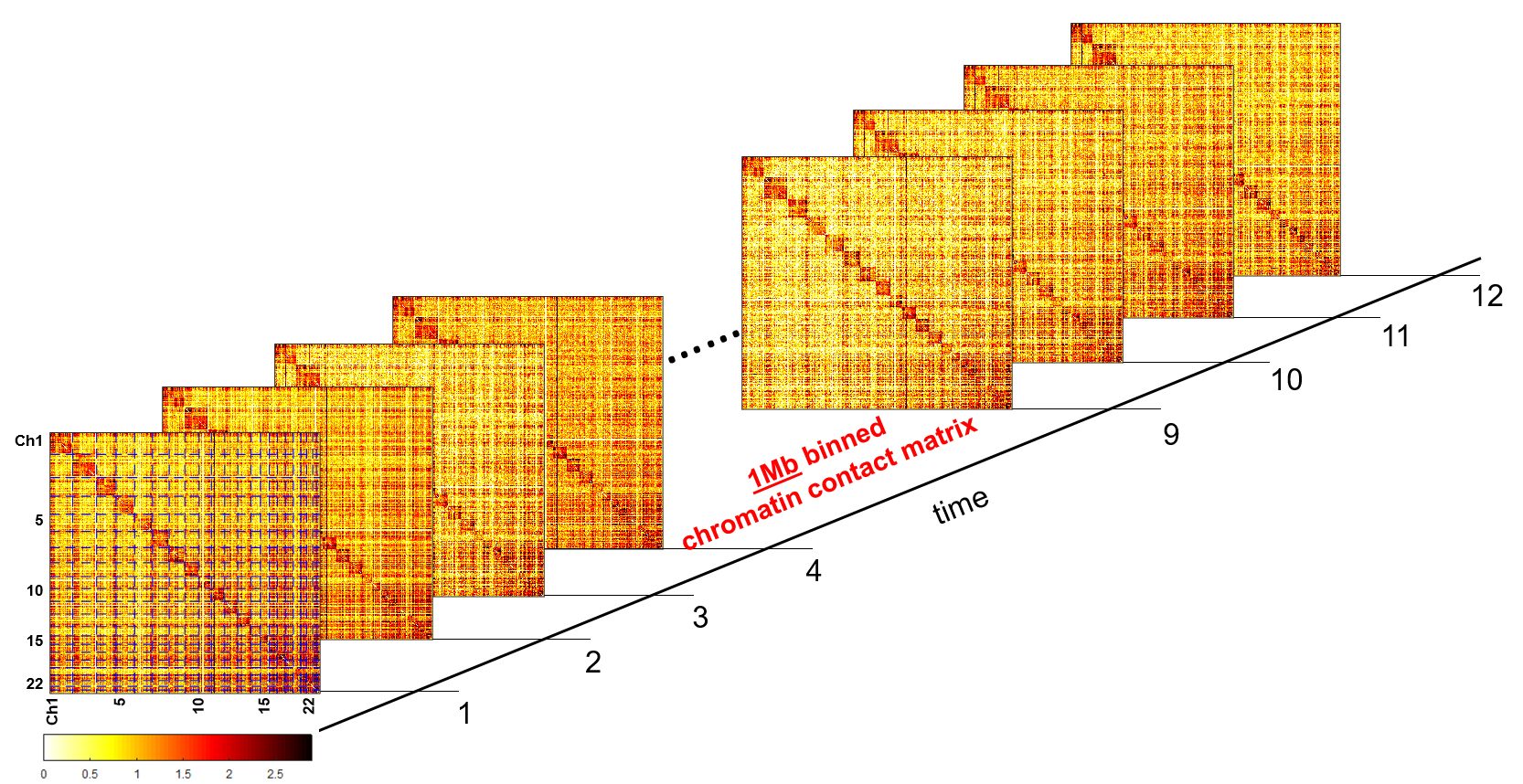}	
	\vspace{0.1mm}
	\caption{Chromatin contact matrix from Hi-C over a time course of 12 samples, which correspond to -48 hour (hr), 0 hr, 8 hr, …, 80 hr over 6 days. }
	\label{Fig_cell_1Mb}
\end{figure*}

\textbf{Rank correlation coefficients.} In addition to PCC, we further use the Spearman's rank correlation coefficient (SRCC) to evaluate the consistency of each method with the anomaly proxy in this task. The results are summarized in Table \ref{table_SRCC}. Similar to the results using PCC, FINGER-JS (Fast) attains the best SRCC among all the compared methods in the four Wikipedia networks. This result again confirms that JS distance via FINGER indeed learns the similar notion of anomaly as indicated by the anomaly proxy.

\section{Addition descriptions for bifurcation detection of cell reprogramming in dynamic genomic networks}

Genome architecture is important in studying cell development, but its dynamics and role in determining cell identity are not well understood. Myogenic differentiation 1 (MYOD1) is a master transcription factor that directly converts human fibroblasts to myogenic cells as studied in \citep{weintraub1989activation,weintraub1993myod}.  Very recently, Liu et al. \citep{liu2018genome}
studied the chromatin contact map (genome-wide structure) through chromosome conformation capture (Hi-C) during the conversion of human fibroblasts to myogenic cells. To understand cell reprogramming, one major question is detecting when the phase transition occurs for cell identity conversion. Liu et al. conducted  experiments and constructed a  1Mb binned chromatin contact matrix (namely, Hi-C matrix) of dimension 2894 over a 6-day time course, leading to 12 sampled measurements. It was found that 
there exists a bifurcation point at the 6th sample (the measurement at 32 hour), suggesting that the cell reprogramming can be interpreted as a genome-wide dynamic system \citep{del2017blueprint} (i.e., a graph sequence) as displayed in Figure \ref{Fig_cell_1Mb}, where the bifurcation occurs when a small structure change made to the cellular system causes a significant system-wide change for genome. 

Liu et al. further used complex graph analysis techniques involving the temporal difference score (TDS) and multiple graph centrality features \citep{CPY16ICASSP} to construct a representative statistic for expressing the states of the studied dynamic genomic contact network as displayed in Figure \ref{Fig_app_cell}, which is used in this paper as the ground-truth statistic for comparing the performance of detecting bifurcation point using  different dissimilarity and distance metrics. In particular, given the TDS of a graph dissimilarity method over measurements, a bifurcation point is defined as the saddle point of the TDS curve excluding the first and last measurements (i.e., $t=1$ and $t=T$). The detected bifurcation point(s) of each method is displayed in Figure \ref{Fig_app_cell}.

\begin{table*}[!t]
	\centering
	\caption{Detection rate on synthesized anomalous events in the dynamic communication networks.}
	\label{table_oregon_full}
	\begin{adjustbox}{max width=1\textwidth}		
		\begin{tabular}{l|lllllllllllll}
			\hline
			\multicolumn{1}{c|}{DoS attack ($X\%$)} & \begin{tabular}[c]{@{}l@{}}FINGER\\ -JS (Fast)\end{tabular} & \begin{tabular}[c]{@{}l@{}}FINGER\\ -JS (Inc.)\end{tabular} & DeltaCon      & RMD           & \begin{tabular}[c]{@{}l@{}}$\lambda$ dist. \\ (Adj.)\end{tabular} & \begin{tabular}[c]{@{}l@{}}$\lambda$ dist. \\ (Lap.)\end{tabular} & GED  & \begin{tabular}[c]{@{}l@{}}VNGE\\ -NL\end{tabular} & \begin{tabular}[c]{@{}l@{}}VNGE\\ -GL\end{tabular} &  VEO  & \begin{tabular}[c]{@{}l@{}}Cosine\\ distance\end{tabular} & \begin{tabular}[c]{@{}l@{}}Bhattacharyya \\ distance\end{tabular} & \begin{tabular}[c]{@{}l@{}}Hellinger\\ distance\end{tabular} \\   \hline
			1 \%                                    & \textbf{24 \%}                                              & 10\%                                                        & 14\%          & 14\%          & 10\%                                                              & \textbf{24\%}                                                     & 14\% & 22\%                                               & 22\%         & 14\%          & 12\%     & 10\%  & 12\%                                                    \\ \hline
			3 \%                                    & \textbf{75\%}                                               & 62\%                                                        & 58\%          & 58\%          & 12\%                                                              & 23\%                                                              & 36\% & 39\%                                               & 39\%     & 36\%  & 35\%     & 14\%                                                                  & 16\%          \\ \hline
			5 \%                                    & \textbf{90\%}                                               & 77\%                                                        & \textbf{90\%} & \textbf{90\%} & 12\%                                                              & 28\%                                                              & 41\% & 67\%                                               & 67\%       & 41\%  & 37\%                                                            & 37\%   & 34\%                  \\ \hline
			10 \%                                   & \textbf{91\%}                                               & \textbf{91\%}                                               & \textbf{91\%} & \textbf{91\%} & \textbf{91\%}                                                     & \textbf{91\%}                                                     & 81\% & \textbf{91\%}                                      & \textbf{91\%}     & 46\%  & 46\%  & 67\%  & 71\%                                                                     \\ \hline
		\end{tabular}
	\end{adjustbox}
\end{table*}

\section{Additional results using VEO as a baseline}
As the VEO score only applies to unweighted undirected graphs, we omit the edge weights in the bifurcation dataset  and find that VEO incorrectly detects graph index 8 as a bifurcation instance. 
In addition, for the synthesized anomaly detection task, VEO only attains $\{46,41,36,14\}$\% detection rate when the DoS attack fraction $X=\{10,5,3,1\}$\%, respectively, as given in Table \ref{table_oregon_full}.

\section{Additional results using degree distribution as dissimilarity metric}
For the synthesized anomalous event detection task, in addition to the dissimilarity metrics in Table \ref{table_oregon}, we also compare the performance of some distance metrics defined on degree distributions -- the cosine distance, the Bhattacharyya distance and the Hellinger distance. We exclude the Kullback-Leibler divergence as the degree distributions of two graphs usually do not have a common support. On the synthesized dataset, Table \ref{table_oregon_full} shows that their performance is not competitive to FINGER and other dissimilarity metrics.



\end{document}